\documentclass[nohyperref]{article}

\usepackage{microtype}
\usepackage{graphicx}
\usepackage{caption}
\usepackage{subcaption}
\usepackage{booktabs} 
\usepackage{bm}
\usepackage{amsmath}
\usepackage{amssymb}
\usepackage{mathtools}
\usepackage{amsthm}
\usepackage{multirow}
\usepackage{hyperref}

\usepackage[accepted]{icml2022}

\usepackage{amsmath}
\usepackage{amssymb}
\usepackage{mathtools}
\usepackage{amsthm}

\usepackage[capitalize,noabbrev]{cleveref}

\theoremstyle{plain}
\newtheorem{theorem}{Theorem}[section]

\theoremstyle{definition}

\theoremstyle{remark}

\usepackage[textsize=tiny]{todonotes}
\usepackage{url}

\icmltitlerunning{Submission and Formatting Instructions for ICML 2022}

\begin{document}
	
	\twocolumn[
	\icmltitle{Robust Meta-learning with Sampling Noise and Label Noise via Eigen-Reptile}

	
	
	\icmlsetsymbol{equal}{*}
	
	\begin{icmlauthorlist}
		\icmlauthor{Dong Chen}{yyy}
		\icmlauthor{Lingfei Wu}{comp}
		\icmlauthor{Siliang Tang}{yyy}
		\icmlauthor{Xiao Yun}{comp}
		\icmlauthor{Bo Long}{comp}
		\icmlauthor{Yueting Zhuang}{yyy}
	\end{icmlauthorlist}
	
	\icmlaffiliation{yyy}{College of Computer Science and Technology, Zhe Jiang University, Hang Zhou, China}
	\icmlaffiliation{comp}{JD.COM Silicon Valley Research Center, 675 E Middlefield Rd, Mountain View, CA 94043 USA}
	
	\icmlcorrespondingauthor{Siliang Tang}{siliang@zju.edu.cn}
	
	\icmlkeywords{Machine Learning, ICML}
	
	\vskip 0.3in
	]
	
	
	
	\printAffiliationsAndNotice{}  

	\begin{abstract}
		Recent years have seen a surge of interest in meta-learning techniques for tackling the few-shot learning (FSL) problem. However, the meta-learner is prone to overfitting since there are only a few available samples, which can be identified as sampling noise on a clean dataset. Moreover, when handling the data with noisy labels, the meta-learner could be extremely sensitive to label noise on a corrupted dataset. To address these two challenges, we present Eigen-Reptile (ER) that updates the meta-parameters with the main direction of historical task-specific parameters to alleviate sampling and label noise. Specifically, the main direction is computed in a fast way, where the scale of the calculated matrix is related to the number of gradient steps instead of the number of parameters. Furthermore, to obtain a more accurate main direction for Eigen-Reptile in the presence of many noisy labels, we further propose Introspective Self-paced Learning (ISPL). We have theoretically and experimentally demonstrated the soundness and effectiveness of the proposed Eigen-Reptile and ISPL. Particularly, our experiments on different tasks show that the proposed method is able to outperform or achieve highly competitive performance compared with other gradient-based methods with or without noisy labels. The code and data for the proposed method are provided for research purposes https://github.com/Anfeather/Eigen-Reptile.
	\end{abstract}
	
	\section{Introduction}
	\label{sec:intro}
	
	Meta-learning, also known as learning to learn, is one of the keys to few-shot learning (FSL) \cite{vinyals2016matching,chi2021meta}, which usually trains meta-parameters as initialization that can fast adapt to new tasks with few samples. 
	One line of the meta-learning is gradient-based methods that optimize meta-parameters by bi-level loop, i.e., inner loop and outer loop, which update task-specific parameters and meta-parameters, respectively. 
	However, fewer samples often lead to a higher risk of overfitting \cite{zintgraf2019fast}, due to the ubiquitous sampling noise and label noise. 
	Particularly, a popular first-order method, Reptile \cite{nichol2018first}, updates the meta-parameters towards the inner loop direction, which is from the current initialization to the last task-specific parameters. 
	
\begin{figure*}[t]
		\centering
		\includegraphics[scale=0.41]{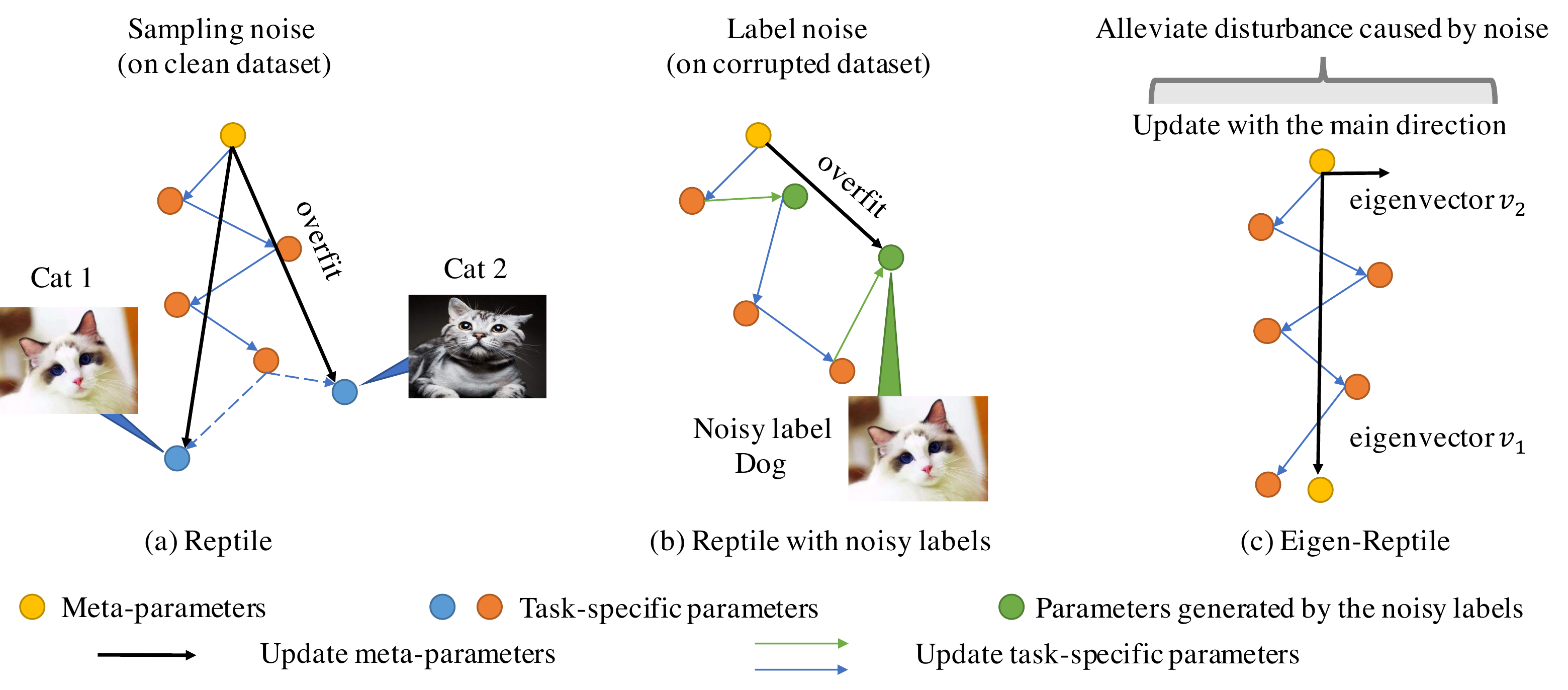}
		\caption{Inner loop steps of Reptile, Eigen-Reptile. Reptile updates meta-parameters towards the last task-specific parameters, which is biased. Eigen-Reptile considers all samples more fair with the main direction of historical task-specific parameters. Note that the main direction is the eigenvector corresponding to the largest eigenvalue. }
		\label{eigen-reptile}
		\vspace{-10pt}
	\end{figure*}

	Sampling noise is a trend for models to overfit the randomly selected samples. As shown in \cref{eigen-reptile}, the model tends to fit the selected sample in every gradient update and results in a high risk of overfitting \cite{zintgraf2019fast}. Specifically, in (a) of \cref{eigen-reptile}, if we sample the cat 1, the meta-learner will have a bias to the left (i.e. overfit cat 1 and the left, bold line is the corresponding update direction) and vice versa, which is an update direction disturbance caused by overfitting sampling noise. 
	Many prior works have proposed different solutions for the aforementioned meta-overfitting problem \cite{zintgraf2019fast}, such as using dropout \cite{bertinetto2018meta, lee2020meta}, and modifying the loss function \cite{jamal2019task} etc. 
	As the disturbance is a random deviation deviated from the unbiased direction according to the selected sample, we regard disturbance as gradient noise \cite{Wu2019On}.
	In other words, we cast the meta-overfitting problem that from sampling noise as disturbance of gradient noise. 
	\cite{neelakantan2015adding} and other works have proved that introducing additional gradient noise can improve the generalization of neural networks with a large number of samples. However, for FSL, there are only a few samples of each task, and the model will overfit the noise \cite{zhang2016understanding} and degrade the model. 
	
	Meta-learner is also inevitably affected by label noise because of the required large number of tasks. Specifically, high-quality manual labeling data is often time-consuming and expensive. And low-cost approaches to collect low-quality annotated data, such as from search engines, will introduce noisy labels. 
	Conceptually, the initialization learned by existing meta-learning algorithms can severely degrade in the presence of noisy labels. As shown in (b) of \cref{eigen-reptile}, noisy labels cause a significant disturbance in the update direction, which is overfitting on corrupted samples \cite{han2020sigua,yu2019does}. 
	Furthermore, conventional algorithms for noisy labels require much more data for each class \cite{yao2018deep,patrini2017making}. Therefore, new methods are needed to alleviate the gradient direction disturbance caused by overfitting label noise.
	
	For the direction disturbance caused by overfitting sampling and label noise, we propose a simple yet effective method, coined Eigen-Reptile (ER), built upon Reptile with a different meta-update direction. In particular, as shown in (c) of \cref{eigen-reptile}, Eigen-Reptile updates the meta-parameters with the main direction of task-specific parameters, which is more unbiased by considering all samples.
	However, it is unrealistic to compute parameters' main direction due to the large scale of neural network parameters. Therefore, we introduce the process of fast computing the main direction into FSL, which computes the eigenvectors of the inner loop step scale matrix instead of the parameter scale matrix.
	Intuitively, noisy labels will degrade the main direction, which in turn degrades the Eigen-Reptile.
	To get a more accurate main direction for Eigen-Reptile, especially when the proportion of noisy labels is high, we propose Introspective Self-paced Learning (ISPL). ISPL combines Self-paced Learning \cite{kumar2010self} with the ensemble idea to discard samples that may degrade the main direction from the dataset. We have theoretically and experimentally demonstrated the soundness and effectiveness of the proposed methods.
	
	Experimental results show that Eigen-Reptile significantly outperforms the baseline, Reptile, by $22.93\%$ and $5.85\%$ on the corrupted and clean dataset, respectively. 
	The main contributions of this paper can be summarized as follows:
	\begin{itemize}
		\item  We cast the meta-overfitting issue as gradient update direction disturbance and discuss the reason for overfitting from the new perspective.
		\item We propose Eigen-Reptile that effectively alleviates disturbance caused by sampling and label noise. Besides, we propose ISPL, which improves the computed main direction in the presence of many noisy labels.
		\item We theoretically verify the effectiveness of the proposed methods for the aforementioned challenges. 
		\item The proposed methods outperform or achieve highly competitive performance compared with recent methods on various benchmark datasets.
	\end{itemize}

	\section{Related Work}
	\noindent\textbf{Meta-Learning with overfitting.}
	There are three main types of meta-learning approaches: metric-based meta-learning approaches \cite{ravi2016optimization,andrychowicz2016learning,santoro2016meta}, model-based meta-learning approaches \cite{vinyals2016matching,koch2015siamese,mordatch2018concept,snell2017prototypical,oreshkin2018tadam} and gradient-based meta-learning approaches \cite{finn2017model,jamal2019task,li2020unsupervised,zintgraf2018caml,li2017meta,rajeswaran2019meta}. In this paper, we focus on gradient-based meta-learning approaches which can be viewed as the bi-level loop. The goal of the outer loop is to update the meta-parameters on a variety of tasks, while task-specific parameters are learned through only a small amount of data in the inner loop. In addition, some algorithms achieve state-of-the-art results by additionally training a model with all classes on meta-training set \cite{yang2020dpgn,hu2020empirical} like transfer learning \cite{zuo2018fuzzy,liu2020towards}; thus, we do not discuss these algorithms in this paper.
	
	Due to too few samples, meta-learner inevitably tends to overfit in FSL \cite{mishra2017simple}. 
	\cite{bertinetto2018meta} find that regularization such as dropout can alleviate meta-overfitting and \cite{yin2019meta} propose meta-regularization on weights; 
	\cite{rajendran2020meta} introduce an information-theoretic framework of meta-augmentation to make meta-learner generalize to new tasks; 
	\cite{ni2021data,yang2021free} improve the performance of meta-learners by data augmentation, which can be combined with various meta-learning algorithms, especially for works \cite{lee2019meta,cao2019theoretical} need more few samples.
	
	\noindent\textbf{Learning with noisy labels.} Learning with noisy labels has been a long-standing problem \cite{frenay2013classification,han2018co,han2018masking}. There are many approaches to solve it, such as studying the denoise loss function \cite{hendrycks2018using,patrini2017making,arazo2019unsupervised}, relabeling \cite{lin2014re}, and so on. 
	Nevertheless, most of these methods require much data for each class. 
	For noisy FSL, a gradient-based meta-learner is trained to optimize an initialization on various tasks with noisy labels. 
	As there are few samples of each class, the traditional algorithms for noisy labels cannot be applied. When the existing gradient-based meta-learning algorithms, such as Reptile, update meta-parameters, they focus on the samples that generate the last gradient step. And these samples may be corrupted, which makes the parameters learned by meta-learner susceptible to noisy labels. To better solve the problem of noisy FSL, we further proposed ISPL based on the idea of Self-paced Learning (SPL) \cite{kumar2010self,khan2011humans,basu2013teaching,tang2012self} to learn more accurate main direction for Eigen-Reptile. ISPL constructs prior models to decide which sample should be discarded when train task-specific models.
	In contrast, the model with SPL learns the samples gradually from easy to complex, and the model itself decides the order, which can improve the robustness like adversarial training \cite{neelakantan2015adding,zhang2020attacks,gao2020maximum,du2021learning}. 

	\section{Preliminaries}
	Gradient-based meta-learning aims to learn meta-parameters $\boldsymbol{\phi}$ as initialization that can adapt to new tasks after a few iterations. The dataset $D$ is usually divided into the meta-training set $D_{meta-train}$ and meta-testing set $D_{meta-test}$, which are used to optimize meta-parameters and evaluate its generalization, respectively. For meta-training, we have tasks $\{\mathcal{T}_i\}^B_{i=1}$ drawn from task distribution $p(\mathcal{T})$, each task has its own train set $D_{train}$ and test set $D_{test}$, and the tasks in  $D_{meta-test}$ are defined in the same way. Note that there are only a small number of samples for each task in FSL. Specifically, the N-way K-shot classification task refers to K examples for each of the N classes. Generally, the number of shots in meta-training should match the one at meta test-time to obtain the best performance \cite{cao2019theoretical}. In this paper, we follow \cite{lee2019meta,cao2019theoretical} to increase the training shots appropriately to get the main direction of individual tasks during meta-training. To minimize the test loss of individual tasks, meta-parameters $\boldsymbol{\phi}$ need to be updated $n$ times to get suitable task-specific parameters $\boldsymbol{\widetilde\phi}$. That is minimizing loss of $N\times K$ samples
	\begin{equation}
	\begin{aligned}
	\min _{\boldsymbol{\phi}} \mathbb{E}_{p(\mathcal{T})}    \left[\frac{1}{N\times K} \sum_{\left(x, y\right) \in \mathcal{D}_{test}} - \log q\left(y \mid x, \boldsymbol{\widetilde\phi}\right)\right]
	\end{aligned}
	\end{equation}
	where $\boldsymbol{\widetilde\phi} = U^n(D_{train}, \boldsymbol{\phi})$, $U^n$ represents $n$ inner loop steps through gradient descent or Adam \cite{kingma2014adam} on batches from $D_{train}$ to update the meta-parameters $\phi$, $q\left( \cdot \right)$ is the predictive distribution. 
	
	When considering updating the meta-parameters in the outer loop, different algorithms have different rules. In the case of Reptile, after $n$ inner loop steps, the meta-parameters will be updated towards the inner loop direction, which is from the current initialization to the last task-specific parameters: 
	\begin{equation}
	\boldsymbol{\phi} \longleftarrow \boldsymbol{\phi} + \beta(\boldsymbol{\widetilde\phi}-\boldsymbol{\phi})
	\end{equation}
	where $\beta$ is a scalar stepsize hyperparameter that controls the update rate of meta-parameters. 
	\section{Method}
	\label{4}
	The proposed Eigen-Reptile alleviates the meta-learner overfitting sampling noise (on a clean dataset) and label noise (on a corrupted dataset) by eliminating the disturbance during gradient update. Furthermore, ISPL improves the performance of Eigen-Reptile by computing a more accurate main direction when there are noisy labels in $D_{meta-train}$.
	
	\subsection{Eigen-Reptile for Clean and Corrupted Dataset}
	To alleviate overfitting sampling and label noise and improve the generalizability of meta-learner, we propose Eigen-Reptile, which updates $d$-dimensional meta-parameters with the main direction of historical task-specific parameters. We train the task-specific model with $n$ inner loop steps (i.e., $n$ gradient updates) that start from the meta-parameters $\boldsymbol{\phi}$. Let i-th column $\boldsymbol{W}_{:,i}\in R^{d\times 1}$ of historical task-specific parameter matrix $\boldsymbol{W}\in R^{d\times n}$ be the parameters after i-th gradient update, i.e., $\boldsymbol{W}_{:,i}=U^i(D_{train}, \boldsymbol{\phi})$. And treat $\boldsymbol{W}_{:,i}$ as a $d$-dimensional parameter point $\boldsymbol{w}_i$ in the parameter space. $\boldsymbol{e} \in R^{d\times 1}$ is a unit vector that represents the main direction of $n$ parameter points in $\boldsymbol{W}$. Intuitively, projecting all parameter points onto $\boldsymbol{e}$ should retain the most information. 
	
	We represent the parameter points by a straight line of the form $\boldsymbol{w} = \boldsymbol{\overline{w}}+l\boldsymbol{e}$, where $\boldsymbol{\overline{w}}$ is the mean point, and $l$ is the signed distance of a point $\boldsymbol{w}$ to $\boldsymbol{\overline{w}}$. Then the loss function 
	$J(l_1,l_2,\cdots,l_n,e) = \sum_{i=1}^{n}{ \parallel \boldsymbol{\overline{w}}+l_{i}\boldsymbol{e}-\boldsymbol{w_i}  \parallel^2 } $.      
	And partially differentiating $J$ with respect to $l_i$, we get $l_i = \boldsymbol{e}^\top (\boldsymbol{w_i} - \boldsymbol{\overline{w}})$. Plugging in this expression for $l_i$ in $J$, 
	\begin{equation}
	\begin{aligned}
	J(\boldsymbol{e}) 
	=-\boldsymbol{e}^\top \boldsymbol{S}\boldsymbol{e}+\sum_{i=1}^{n}{\parallel \boldsymbol{w_i}-\boldsymbol{\overline{w}} \parallel^2}
	\label{5}
	\end{aligned}
	\end{equation}
	where $\boldsymbol{S}=\sum_{i=1}^{n}{ (\boldsymbol{w_i}-\boldsymbol{\overline{w}})(\boldsymbol{w_i}-\boldsymbol{\overline{w}})^\top }$ is a scatter matrix. According to \cref{5}, minimizing $J$ is equivalent to maximizing $\boldsymbol{e}^\top \boldsymbol{S}\boldsymbol{e}$. Note that $\boldsymbol{e}$ needs to be roughly consistent with the gradient update direction $\boldsymbol{\overline{V}}$ in the inner loop, as the computed $\boldsymbol{e}$ without $\pm$. Therefore, we add a linear constraint $\boldsymbol{\overline{V}}\boldsymbol{e}>0$ and use Lagrange multiplier method as
	\begin{tiny}
		\begin{equation}
		\begin{split}
		\label{FX}
		\max  \boldsymbol{e}^\top \boldsymbol{S}\boldsymbol{e} 
		\qquad	\textrm{s.t.}
		\begin{cases}
		\text{ $\boldsymbol{\overline{V}}\boldsymbol{e}>0$} \\
		\text{$\boldsymbol{e}^\top \boldsymbol{e}=1$}
		\end{cases},  
		\text{where}\quad \boldsymbol{\overline{V}} = \frac{1}{\lfloor n/2\rfloor}\sum_{i=1}^{\lfloor n/2\rfloor}{\boldsymbol{w}_{n-i+1}-\boldsymbol{w}_{i}}
		\end{split}
		\end{equation}
	\end{tiny}
	We get the objective function 
	\begin{equation}
	\begin{aligned}
	&g(\mu,\boldsymbol{e},\lambda,\eta)=\boldsymbol{e}^\top \boldsymbol{S}\boldsymbol{e} - \lambda (\boldsymbol{e}^\top \boldsymbol{e}-1) + \mu(-\boldsymbol{\overline{V}}\boldsymbol{e}+\eta^2) ,\\
	&\textrm{where}\quad \lambda \not=0, \mu \ge0
	\label{object}
	\end{aligned}
	\end{equation}
	then partially differentiating $g$ in \cref{object} with respect to $\mu,e,\lambda,\eta$,
	\begin{align}
	\left\{
	\begin{aligned}
	-\boldsymbol{\overline{V}}\boldsymbol{e}+\eta^2&=0 \\
	2\boldsymbol{S}\boldsymbol{e}-2\lambda \boldsymbol{e}-\mu\boldsymbol{\overline{V}}&=0 \\
	\boldsymbol{e}^\top \boldsymbol{e}-1&=0\\
	2\mu\eta &=0
	\end{aligned}
	\right.
	\label{KKT}
	\end{align}
	According to \cref{KKT}, if $\eta=0$, then $\boldsymbol{\overline{V}}$ and $\boldsymbol{e}$ are orthogonal, which obviously does not meet our expectations. So we get $\eta\not=0$, and $\mu=0$, then $\boldsymbol{S}\boldsymbol{e}=\lambda \boldsymbol{e}$. We can see $\boldsymbol{e}$ is the eigenvector of $\boldsymbol{S}$ corresponding to the largest eigenvalue $\lambda$, which is the required main direction. It should be noted that even if $\boldsymbol{\overline{V}}$ is not directly related to $\boldsymbol{e}$, in Eigen-Reptile, the linear constraint $\boldsymbol{\overline{V}}\boldsymbol{e}>0$ in \cref{FX} and line 23 of \cref{A1} must be retained as it determines the update direction of the outer-loop. Otherwise, the algorithm will not converge, which has been proven empirically.
	
	A concerned question about $\boldsymbol{S}\boldsymbol{e}=\lambda \boldsymbol{e}$ is that the scatter matrix $\boldsymbol{S} \in R^{d\times d}$ grows quadratically with the number of parameters $d$. As the large number of of parameters typically used in neural networks, computing eigenvalues and eigenvectors of $\boldsymbol{S}$ could come at a prohibitive cost (the worst-case complexity is $\mathcal{O}\left(\mathrm{d}^{3}\right)$ ). To avoid calculating the eigenvectors of $\boldsymbol{S}$ directly, we focus on $\boldsymbol{W}^\top \boldsymbol{W}$ (centralize $\boldsymbol{W}$ by subtracting the mean $\boldsymbol{\overline{w}}$, and the scatter matrix $\boldsymbol{S}=\boldsymbol{W}\boldsymbol{W}^\top$). As $\boldsymbol{W}^\top \boldsymbol{W}\boldsymbol{\widehat{e}}=\widehat{\lambda}\boldsymbol{\widehat{e}}$, multiply both sides of the equation with $\boldsymbol{W}$,
	\begin{align}
	\boldsymbol{WW}^\top \underbrace{\boldsymbol{W}\boldsymbol{\widehat{e}}}_{\boldsymbol{e}}=\underbrace{\widehat{\lambda}}_{\lambda}\underbrace{\boldsymbol{W}\boldsymbol{\widehat{e}}}_{\boldsymbol{e}}
	\label{8} 
	\end{align}
	It can be found from \cref{8} that $\boldsymbol{W}^\top \boldsymbol{W} \in R^{n\times n}$ and $\boldsymbol{W} \boldsymbol{W}^\top \in R^{d\times d}$ have the same eigenvalue, $\lambda=\widehat{\lambda} $. Furthermore, we get the eigenvector of $\boldsymbol{W} \boldsymbol{W}^\top$ as $\boldsymbol{e}=\boldsymbol{W}\boldsymbol{\widehat{e}}$. The main advantage of \cref{8} is that the intermediate matrix $\boldsymbol{W}^\top \boldsymbol{W}$ now grows quadratically with the inner loop steps.
	As we are interested in FSL, $n$ is very small. It will be much easier to compute the eigenvector $\boldsymbol{\widehat{e}}$ of $\boldsymbol{W}^\top \boldsymbol{W}$. Then we get the eigenvector $\boldsymbol{e}$ of $\boldsymbol{W}\boldsymbol{W}^\top$ based on $\boldsymbol{\widehat{e}}$. Moreover, we project parameter update vectors $\boldsymbol{w}_{i+1}-\boldsymbol{w}_i, i=1,2,\cdots,n-1$ on $\boldsymbol{e}$ to get the corresponding update stepsize $\nu$,  so meta-parameters $\boldsymbol{\phi}$ can be updated as
	\begin{tiny}
		\begin{equation}
		\begin{aligned}
		\boldsymbol{\phi} \longleftarrow \boldsymbol{\phi} + \beta\nu\zeta \boldsymbol{e}, \textrm{where}\quad \zeta = \frac{\lambda}{\sum_{m=1}^{n} {\lambda_m}},\quad \nu=\sum_{i=1}^{n-1}(\boldsymbol{w}_{i+1}-\boldsymbol{w}_{i})\boldsymbol{e}
		\end{aligned}
		\end{equation}
	\end{tiny}
	where $\beta$ is a scalar stepsize hyperparameter that controls the update rate of meta-parameters, $\zeta$ is the proportion of the largest eigenvalue to the sum of all eigenvalues. The larger the value of $\zeta$, the more accurate the meta-parameter update direction. The Eigen-Reptile algorithm is summarized in \cref{A1}. 
	\subsection{Analysis of Eigen-Reptile}
	To illustrate the validity of Eigen-Reptile for alleviating overfitting sampling noise, we present \cref{TT1} (the gradient disturbances generated by overfitting sampling noise are slightly random deviations according to the selected samples, which can be regarded as gradient noise \cite{Wu2019On}).
	\begin{theorem}
		Assume that the gradient noise variable $x$ follows Gaussian distribution \cite{hu2017diffusion,jastrzkebski2017three,mandt2016variational}, i.e., $x\sim\mathrm{N}\left(0, \sigma^2\right)$. Moreover, $x$ and neural network parameter variable are assumed to be uncorrelated. The observed covariance matrix $\boldsymbol{C}$ equals noiseless covariance matrix $\boldsymbol{C}_t$ plus gradient noise covariance matrix $\boldsymbol{C}_x$. Then, we get
		\begin{equation}
		\centering
		\begin{aligned}
		\boldsymbol{C} &=\frac{1}{n-1} \boldsymbol{S} =
		\boldsymbol{C}_t+\boldsymbol{C}_x =\boldsymbol{P}_t(\Lambda_t  + \boldsymbol{\Lambda}_x )\boldsymbol{P}_t^\top\\
		&=\boldsymbol{P}_t(\boldsymbol{\Lambda}_t  + \sigma^2 \boldsymbol{I} )\boldsymbol{P}_t^\top=\boldsymbol{P}_t\Lambda \boldsymbol{P}_t^\top=\boldsymbol{P}\Lambda \boldsymbol{P}^\top
		\label{ETT1}
		\end{aligned}
		\end{equation}
		where $\boldsymbol{P}_t$ and $\boldsymbol{P}$ are the orthonormal eigenvector matrices of $\boldsymbol{C}_t$ and $\boldsymbol{C}$ respectively, $\boldsymbol{\Lambda}_t$ and $\boldsymbol{\Lambda}$ are the corresponding diagonal eigenvalue matrices, and $\boldsymbol{I}$ is an identity matrix. It can be seen from \cref{ETT1} that $\boldsymbol{C}$ and $\boldsymbol{C}_t$ has the same eigenvectors. 
		We defer the proof to the \textbf{\textit{Appendix}} \ref{AppendB}.
		\label{TT1}
	\end{theorem}
	\cref{TT1} shows that eigenvectors are not affected by gradient noise. Therefore, Eigen-Reptile can find a more generalizable starting point for new tasks without overfitting sampling noise (on the clean dataset). As for label noise (on the corrupted dataset), the analysis is shown in  \textbf{\textit{Appendix}} \ref{AppendD}. 
	We also show the complexity analysis in \textbf{\textit{Appendix}} \ref{AppendC}, which illustrates that Reptile and Eigen-Reptile are the same in spatial complexity and time complexity.
	
	\begin{algorithm}[!h]
		\caption{Eigen-Reptile}
		\textbf{Input}: Distribution over tasks $P(\mathcal{T})$, outer step size $\beta$.
		\begin{algorithmic}[1]
			\STATE Initialize meta-parameters $\boldsymbol{\phi}$
			\WHILE{not converged}
			\STATE $\boldsymbol{W}= [\ ],\nu=0$ 
			\STATE Sample batch of tasks $\{\mathcal{T}_i\}^B_{i=1} \sim P(\mathcal{T})$
			\FOR{each task $\mathcal{T}_i$}
			\STATE $\boldsymbol{\phi}_i = \boldsymbol{\phi}$
			\STATE Sample train set $D_{train}$ of $\mathcal{T}_i$ 
			\FOR{$j = 1,2,3,...,n$}
			\STATE $\boldsymbol{\phi}_i^j = U^j(D_{train},\boldsymbol{\phi}_i)$
			\STATE $\boldsymbol{W}$ appends $\boldsymbol{W}_{:,j}=flatten(\boldsymbol{\phi}_i^j)$
			\ENDFOR
			\STATE Mean centering, $\boldsymbol{W}=\boldsymbol{W}-\overline{\boldsymbol{w}}, \quad \overline{\boldsymbol{w}}\in R^{d \times 1}$
			\STATE Compute matrix $\widehat{\boldsymbol{\Lambda}}$ and eigenvector matrix $\widehat{\boldsymbol{P}}$ of scatter matrix $\boldsymbol{W}^\top \boldsymbol{W}$
			\STATE Eigenvalues $\lambda_1>\lambda_2>\cdots>\lambda_n$ in $\widehat{\boldsymbol{\Lambda}}$
			\STATE Compute matrix of $\boldsymbol{W} \boldsymbol{W}^\top$, $\boldsymbol{P}=\boldsymbol{W}\widehat{\boldsymbol{P}}$
			\STATE Let the eigenvector corresponding to $\lambda_1$ be a unit vector, $\parallel \boldsymbol{e}_i^1\parallel_2^2 =1$			
			\FOR{$j = 1,2,3,...,n-1$}
			\STATE $\nu=\nu+(\boldsymbol{W}_{:,j+1}-\boldsymbol{W}_{:,j})\boldsymbol{e}_i^1$
			\ENDFOR
			\STATE $\boldsymbol{e}_i^1=\frac{\lambda_1}{\sum_{m=1}^{n} {\lambda_m}} \times \boldsymbol{e}_i^1$	
			\STATE Calculate the approximate direction of task-specific gradient update $\overline{\boldsymbol{V}}$: 
			\STATE $\overline{\boldsymbol{V}} = \frac{1}{\lfloor n/2\rfloor}\sum_{i=1}^{\lfloor n/2\rfloor}{\boldsymbol{W}_{:,n-i+1}-\boldsymbol{W}_{:,i}}$
			\IF{$\boldsymbol{e}_i^1 \cdot \overline{\boldsymbol{V}}<0$}
			\STATE	$\boldsymbol{e}_i^1= -\boldsymbol{e}_i^1$
			\ENDIF
			\ENDFOR
			\STATE Average the main directions to get \\
			\qquad $\tilde{\boldsymbol{e}} = (1/B) \sum_{i=1}^{B}{\boldsymbol{e}_i^1}$
			\STATE Update meta-parameters $\boldsymbol{\phi} \longleftarrow \boldsymbol{\phi} + \beta \times \nu/B \times \tilde{\boldsymbol{e}}$
			\ENDWHILE
		\end{algorithmic}
		\label{A1}
	\end{algorithm}

	\subsection{The Introspective Self-paced Learning for More Accurate Main Direction}
	
	\begin{figure}[h]
		\centering
		\includegraphics[width=0.32\textwidth]{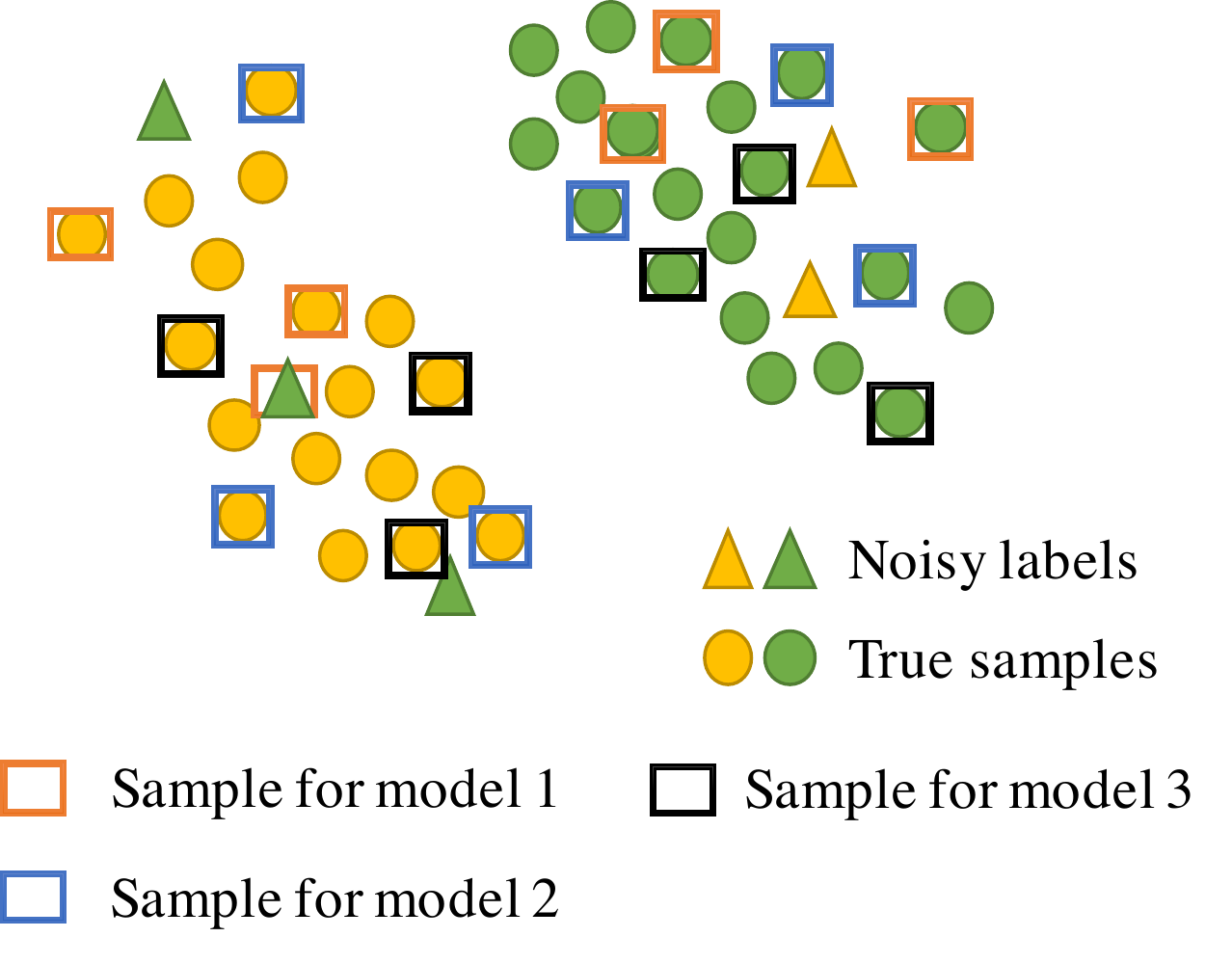}
		\vspace{-5pt}
		\caption{ Randomly sample examples to build prior models. }
		\label{sample_models}
		\vspace{-5pt}
	\end{figure}

	As shown in \textbf{\textit{Appendix}} \ref{AppendD}, Eigen-Reptile addresses the noisy FSL problem by separating noisy information. However, with the increase of noise ratio, the eigenvector will gradually become invalid. To get a more accurate eigenvector, we propose Introspective Self-paced Learning (ISPL).
	
	Self-paced learning (SPL) learns the samples from low losses to high losses, which is proven beneficial in achieving a better generalization result \cite{khan2011humans,basu2013teaching,tang2012self}. Besides, some previous work \cite{zhu2019self} solve the problem of traditional noisy labels by SPL. Nevertheless, in a meta-learning setting, a meta-learner is trained on various tasks; the initial model may have lower losses for trained classes and higher losses for unseen classes or noisy samples. For this reason, we cannot train the task-specific model in the same way as SPL to solve the noisy FSL problem. To this end, we propose an improved SPL algorithm to help Eigen-Reptile achieve better performance for the problem of the noisy labels. As shown in \cref{sample_models}, even though the two categories of the yellow and green show an excellent distribution that can be well separated, some samples are marked wrong. Then, we build three prior models. Specifically, we randomly sample three times, and model 1 is trained with a corrupted label. Due to different samples learned by prior models, building multiple models to vote on the data will obtain more accurate losses, which is a kind of ensemble learning. 
	Moreover, samples with losses above a certain threshold will be discarded. Furthermore, we imitate SPL to add the hidden variable $v = 0$ or $1$ that is decided by $Q$ prior models before the loss of each sample to control whether the sample should be abandoned. And we get the task-specific loss as
	\begin{equation}
	\begin{aligned}
	& L_{ISPL}\left(\boldsymbol{\phi}, \boldsymbol{v}\right)=  \sum_{i=1}^{h} v_{i} L\left(x_{i}, y_{i}, \boldsymbol{\phi}\right), \\
	& \textrm{where} \quad	v_i=  \arg \min _{{v_i}}  \frac{v_i}{Q}\sum_{j=1}^{Q} L_j\left(x_{i}, y_{i}, \boldsymbol{\phi}_j\right)-\gamma v_{i}
	\label{EQ:ISPL}
	\end{aligned}
	\end{equation}
	where $h$ is the number of samples $x$ from dataset $D_{train}$, $y$ is label, $\gamma$ is the sample selection parameter, which gradually decreases, parameter of model $j$ is  $\boldsymbol{\phi}_j=U^n(D_j,\boldsymbol{\phi}), D_j\in D_{train}$. Note that we update the meta-parameters with the model trained on $h$ samples from $D_{train}$. The objective of \cref{EQ:ISPL} is choosing samples whose summary loss is lower than $\gamma$, as the hidden variable $v=1$, and vice versa.
	ISPL is summarized in \textit{\textbf{Appendix}} \ref{AppendA}. 
	\subsection{Analysis of ISPL}
	Intuitively, it is difficult to say whether discarding high-loss samples containing correct and corrupted samples will improve the accuracy of eigenvector, so we prove the effectiveness of ISPL by \cref{TT2}.
	\begin{figure*}[t!]
		\centering
		\begin{subfigure}[b]{0.17\textwidth}
			\centering
			\includegraphics[width=3cm,height=1.2cm]{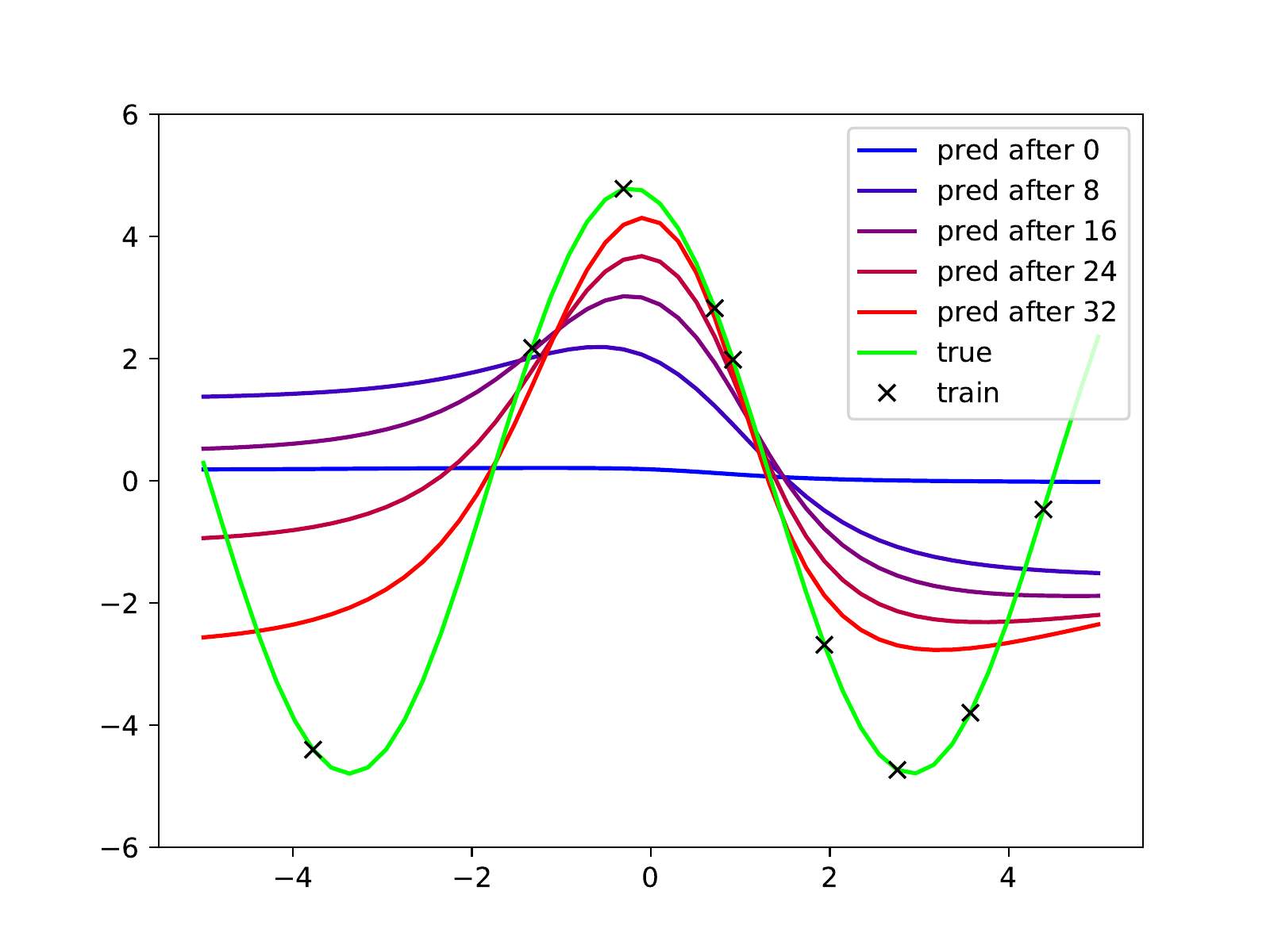}
			\caption{ER iteration 1}
		\end{subfigure}
		\begin{subfigure}[b]{0.17\textwidth}
			\centering
			\includegraphics[width=3cm,height=1.2cm]{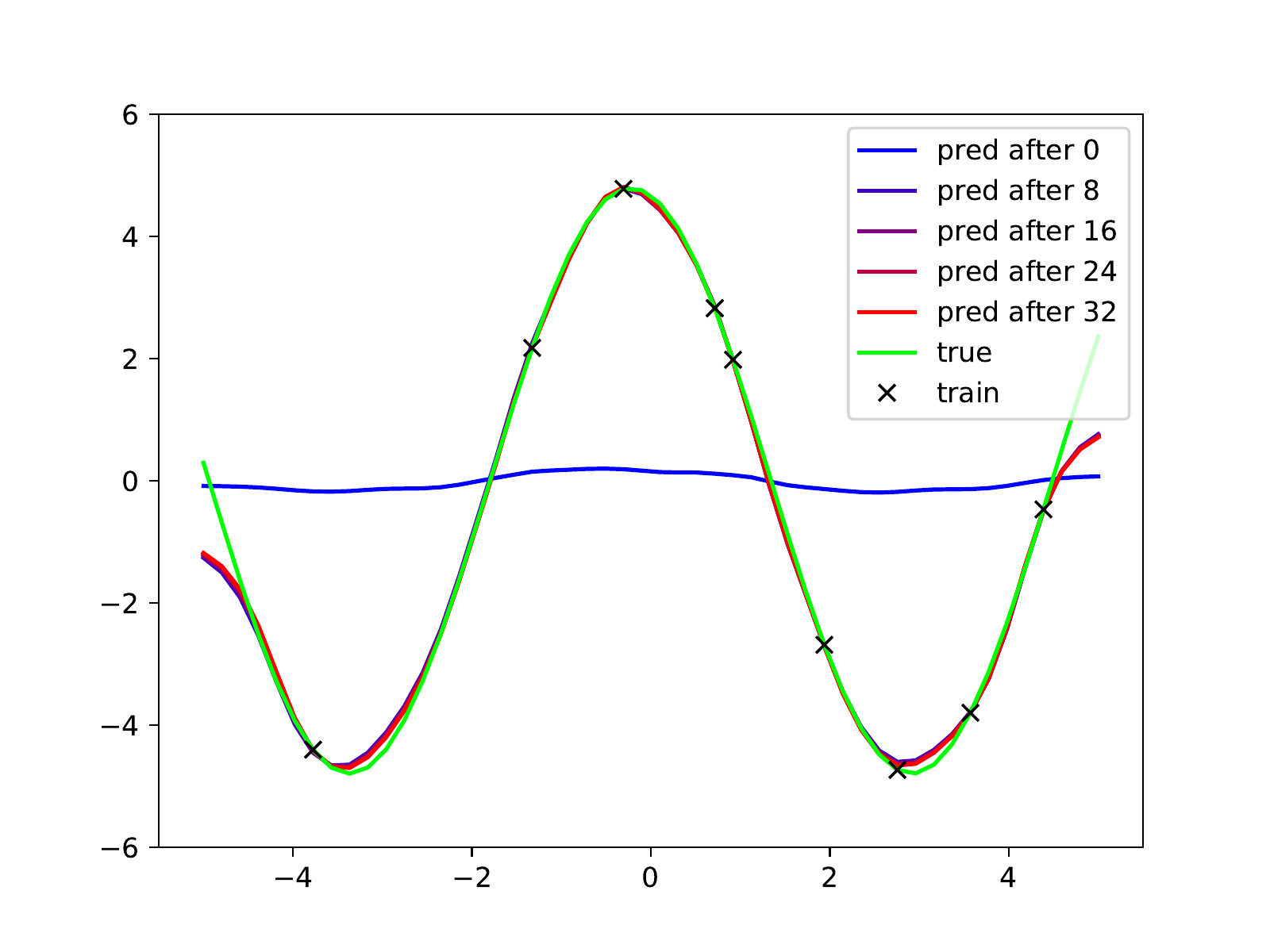}
			\caption{ER iteration 10000}
		\end{subfigure}
		\begin{subfigure}[b]{0.17\textwidth}
			\centering
			\includegraphics[width=3cm,height=1.2cm]{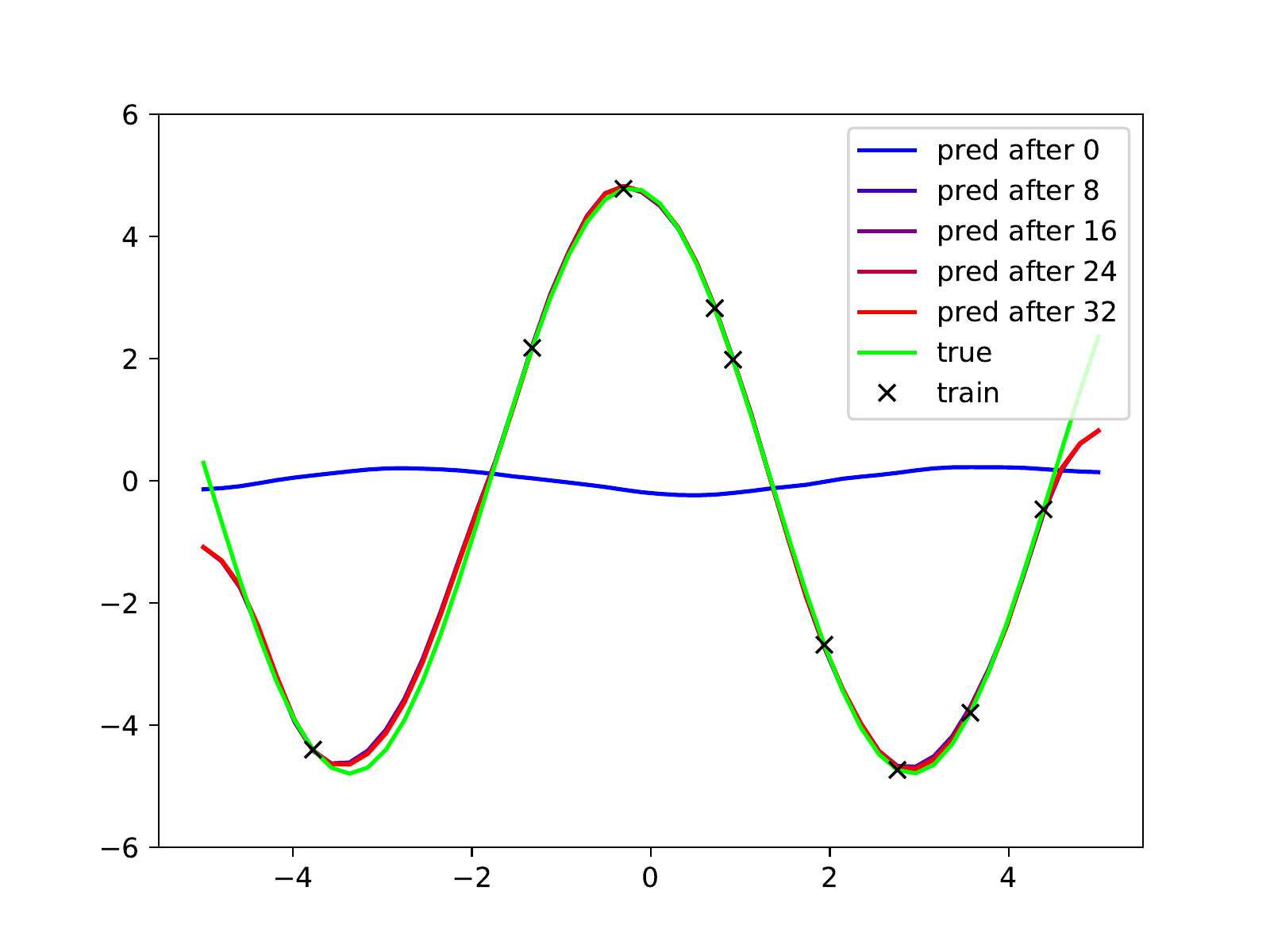}
			\caption{ER iteration 15000}
		\end{subfigure}
		\begin{subfigure}[b]{0.17\textwidth}
			\centering
			\includegraphics[width=3cm,height=1.2cm]{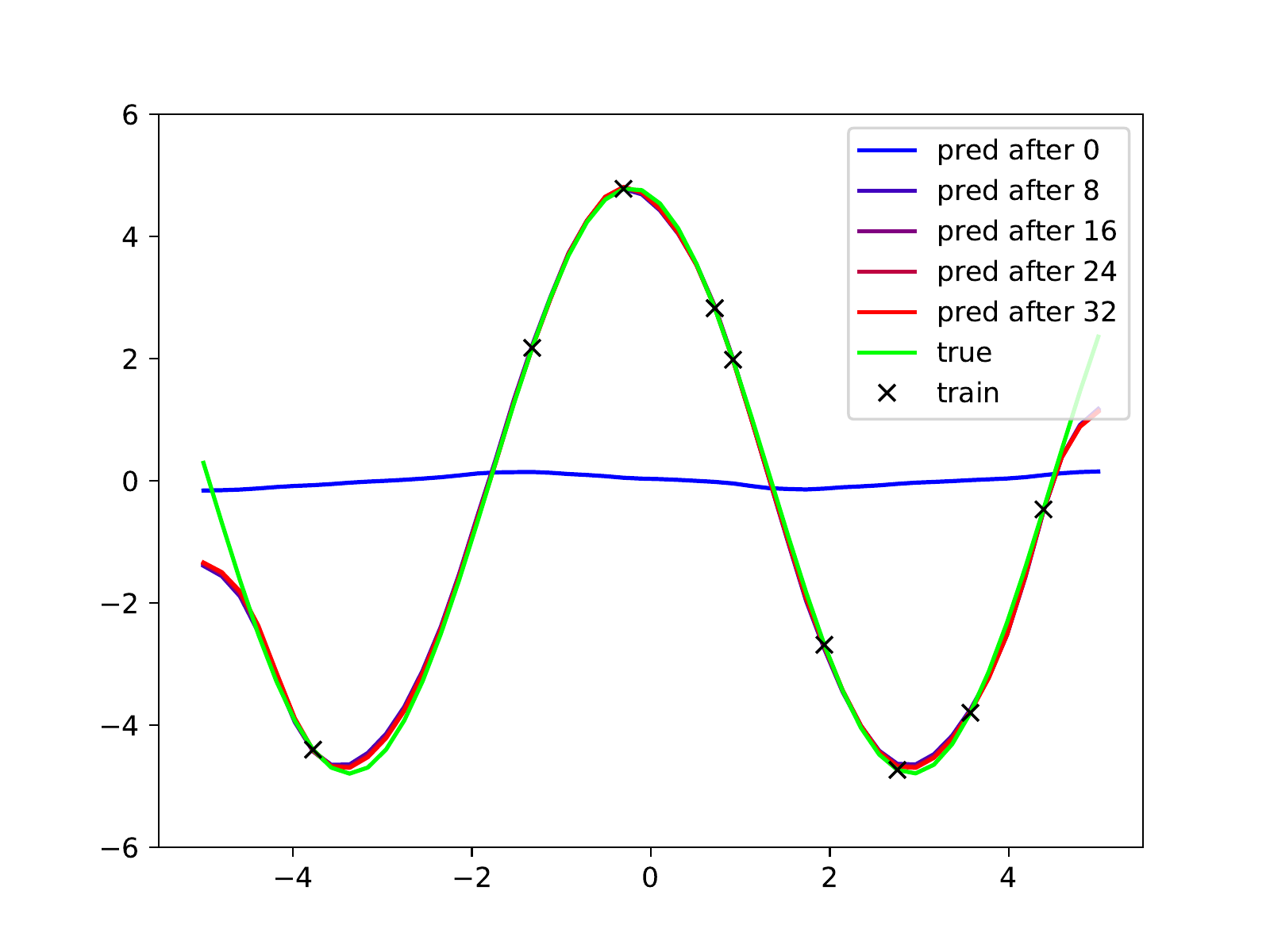}
			\caption{ER iteration 20000}
		\end{subfigure}
		\begin{subfigure}[b]{0.17\textwidth}
			\centering
			\includegraphics[width=3cm,height=1.2cm]{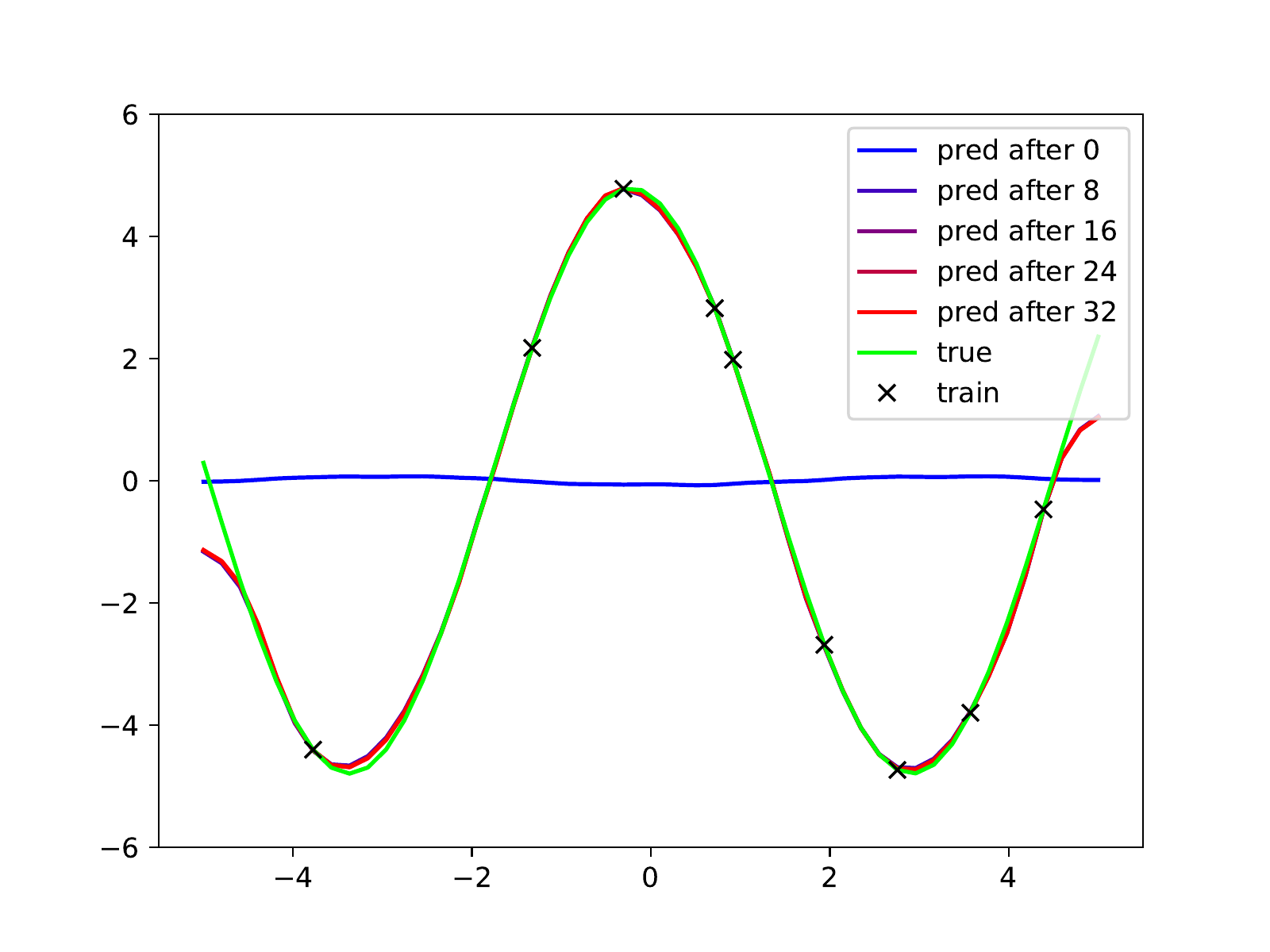}
			\caption{ER iteration 30000}
		\end{subfigure}
		
		\begin{subfigure}[b]{0.17\textwidth}
			\centering
			\includegraphics[width=3cm,height=1.2cm]{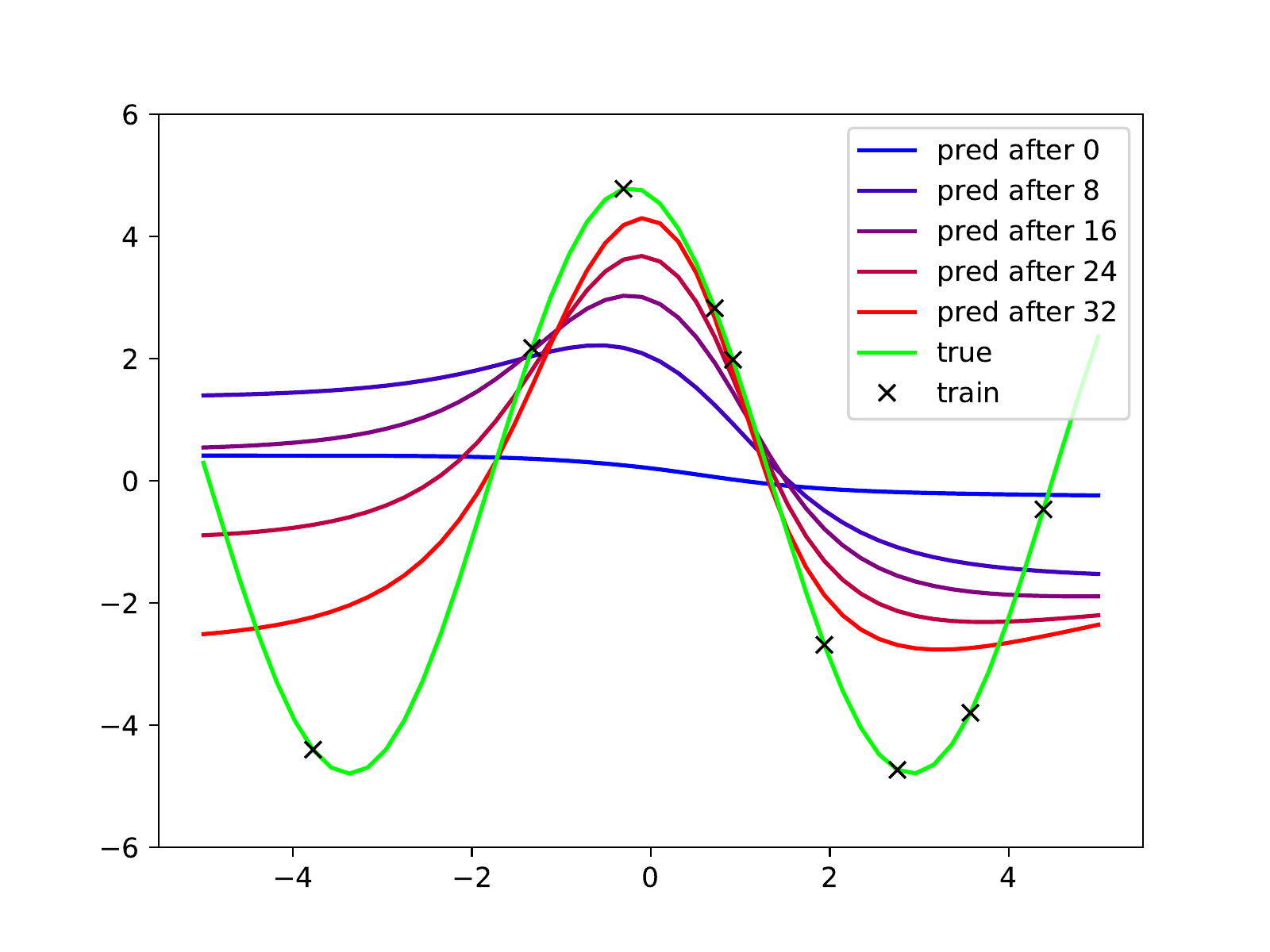}
			\caption{R iteration 1}
		\end{subfigure}
		\begin{subfigure}[b]{0.17\textwidth}
			\centering
			\includegraphics[width=3cm,height=1.2cm]{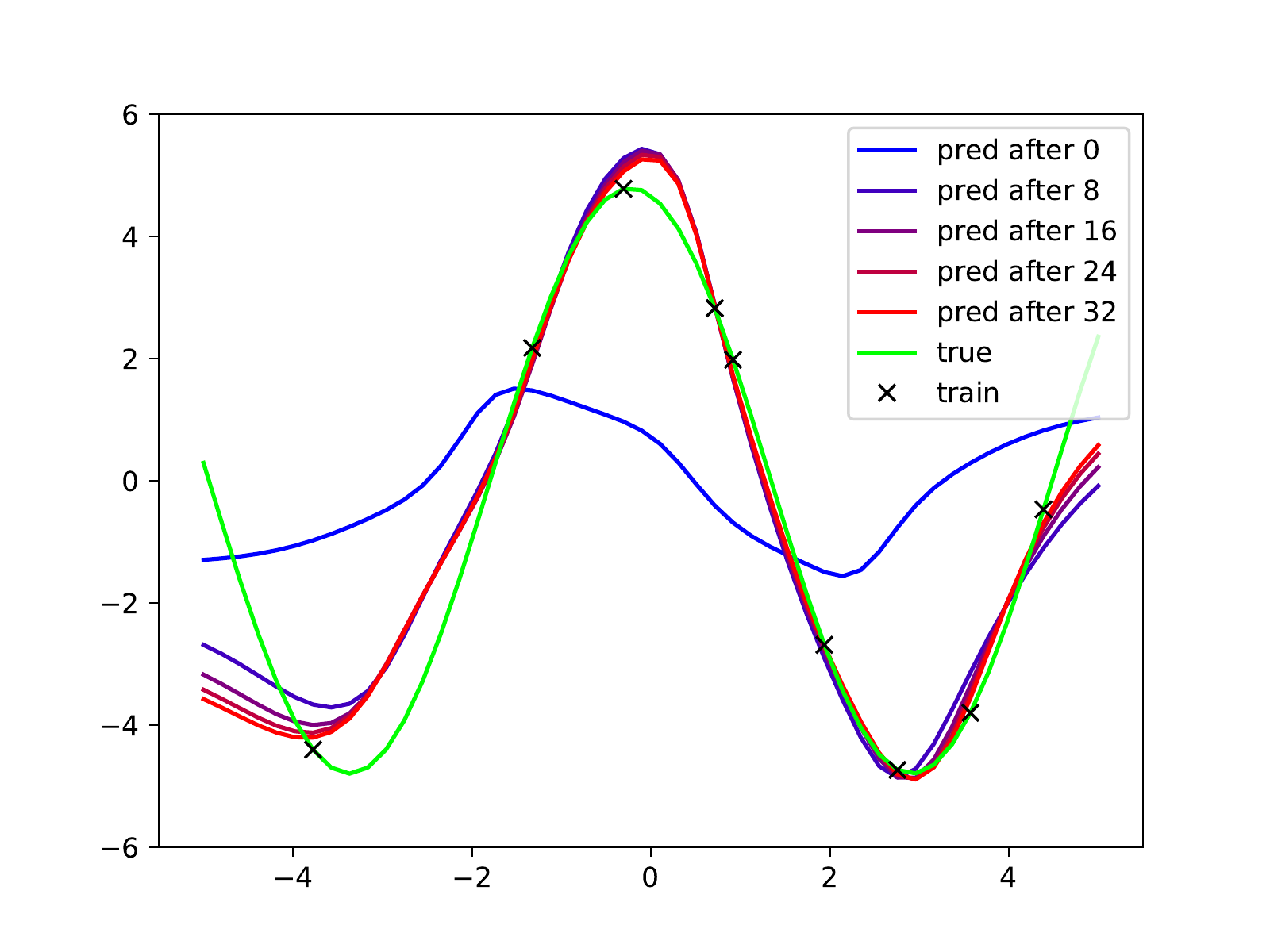}
			\caption{R iteration 10000}
		\end{subfigure}
		\begin{subfigure}[b]{0.17\textwidth}
			\centering
			\includegraphics[width=3cm,height=1.2cm]{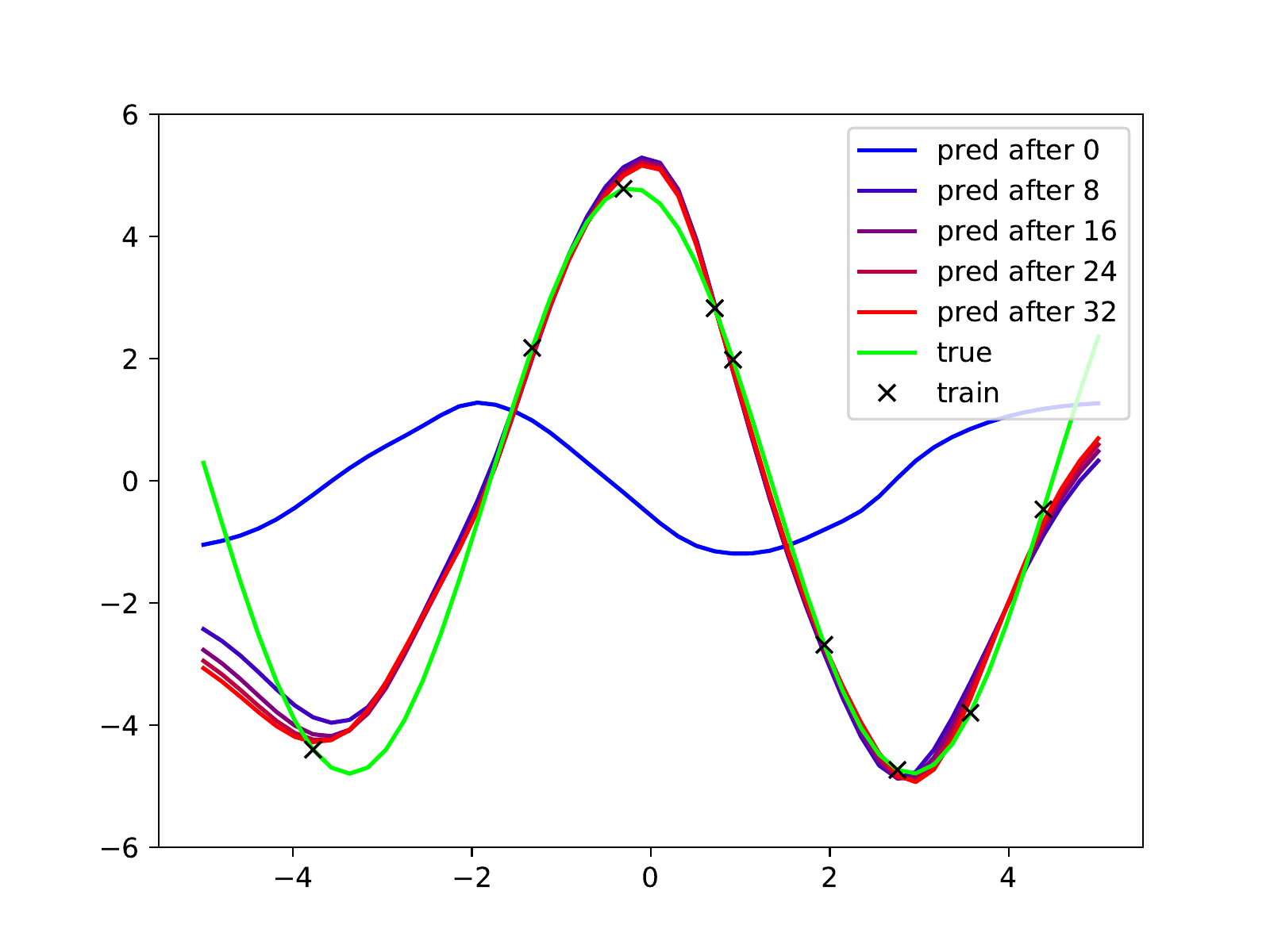}
			\caption{R iteration 15000}
		\end{subfigure}
		\begin{subfigure}[b]{0.17\textwidth}
			\centering
			\includegraphics[width=3cm,height=1.2cm]{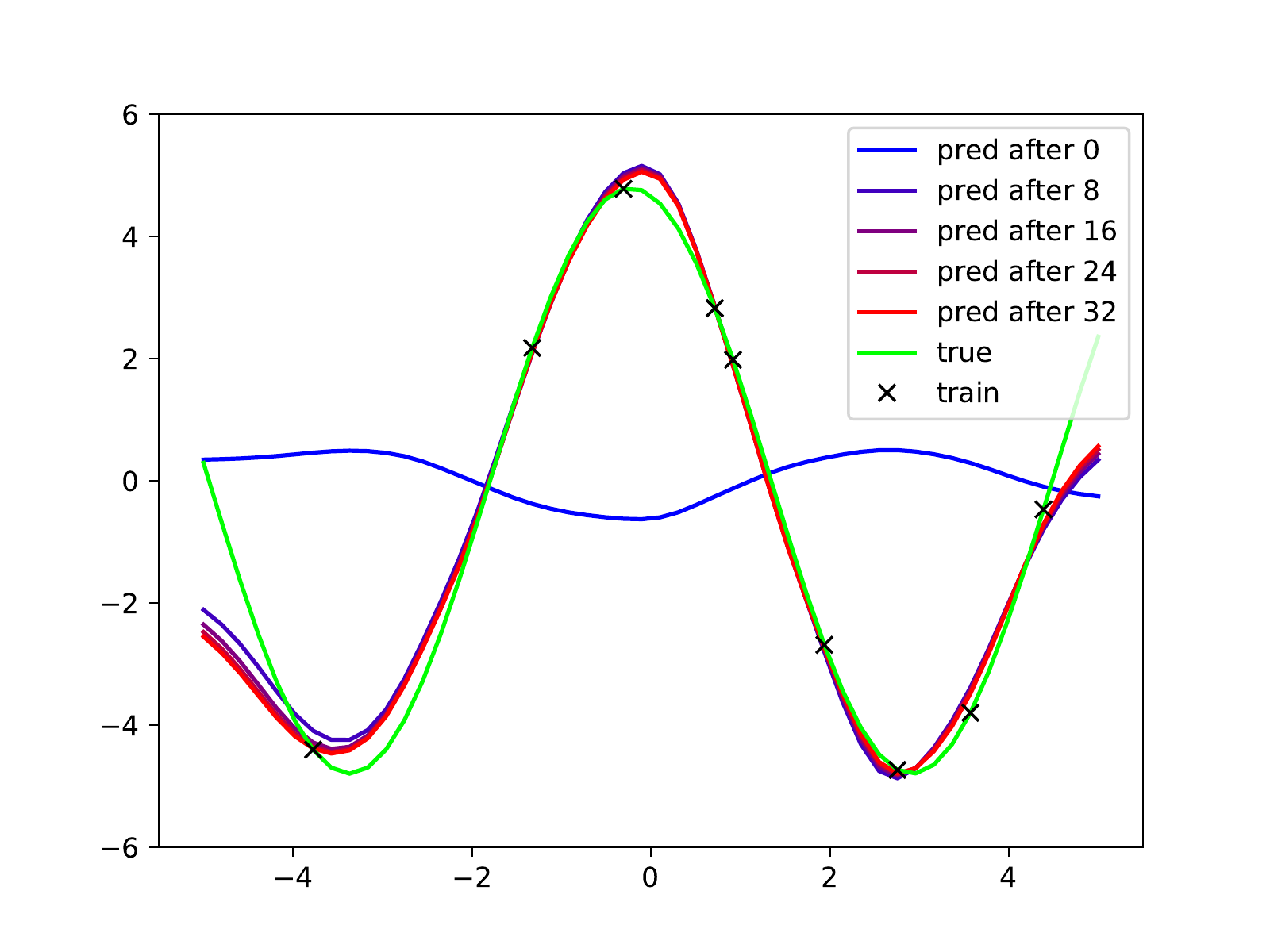}
			\caption{R iteration 20000}
		\end{subfigure}
		\begin{subfigure}[b]{0.17\textwidth}
			\centering
			\includegraphics[width=3cm,height=1.2cm]{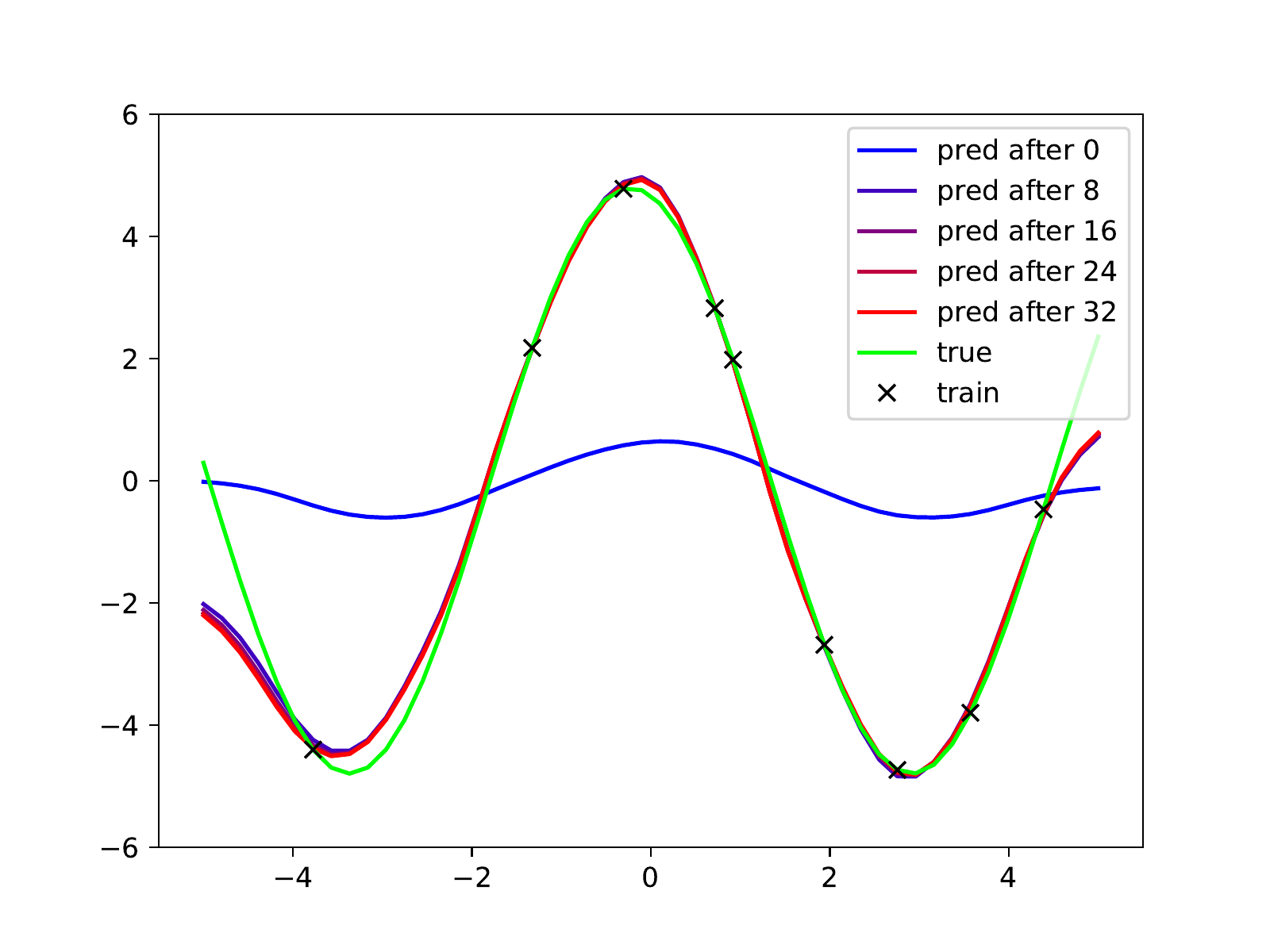}
			\caption{R iteration 30000}
		\end{subfigure}
		\vspace{-5pt}
		\caption{Eigen-Reptile (ER) and Reptile (R) training process on the regression toy test. (a), (b), (c), (d), (e) and (f), (g), (h), (i), (j) show that after the gradient update 0, 8, 16, 24, 32 times based on initialization learned by Eigen-Reptile and Reptile respectively. }
		\label{img4}
		\vspace{-10pt}
	\end{figure*}
	
	\begin{theorem}
		Let $\boldsymbol{W}_o$ be the parameter matrix only generated by the corrupted samples.
		Compute the eigenvalues and eigenvectors of the observed expected parameter matrix
		\begin{equation}
		\begin{aligned}
		&\frac{1}{\lambda} \mathbb{E}(\boldsymbol{C}_{tr})\boldsymbol{e}\\&=
		\boldsymbol{P}_{o}(\boldsymbol{I}-\frac{\boldsymbol{\Lambda}_{o}}{\lambda})\boldsymbol{P}_{o}^\top \boldsymbol{e} \\
		&\approx  \boldsymbol{P}_{o}(\boldsymbol{I}-\frac{\lambda_{o}}{\lambda}\boldsymbol{I})\boldsymbol{P}_{o}^\top \boldsymbol{e}
		>\boldsymbol{P}_{o}(\boldsymbol{I}-\frac{\lambda_{o}-\xi}{\lambda-\xi}\boldsymbol{I})\boldsymbol{P}_{o}^\top \boldsymbol{e}
		\label{PP12}
		\end{aligned}
		\end{equation}
		where $\boldsymbol{C}_{tr}$ is the covariance matrix generated by clean samples, $\lambda$ is the observed largest eigenvalue, $\lambda_{o}$ is the largest eigenvalue in the corrupted diagonal eigenvalue matrix $\boldsymbol{\Lambda}_{o}$, $\boldsymbol{P}_o$ is the orthonormal eigenvector matrix of corrupted covariance matrix. According to \cref{PP12}, if $\lambda_{o} /\lambda$ is smaller, the observed eigenvector $e$ is more accurate. Assume that the discarded high loss samples have the same contributions $\xi$ to $\lambda$ and $\lambda_{o}$, representing the observed and corrupted main directional variance, respectively. Note that these two kinds of data have the same effect on the gradient updating of the model, as they all generate high loss (neither new clean samples from new tasks nor corrupted samples are not familiar to the model). 
		Furthermore, it is easy to find that $(\lambda_{o}-\xi) /(\lambda-\xi)$ is smaller than $\lambda_{o} /\lambda$. We defer the proof to the \textbf{\textit{Appendix}} \ref{AppendE}.
		\label{TT2}
	\end{theorem}
	
	\cref{TT2} shows that discard high loss samples can improve the accuracy of the observed eigenvector learned with corrupted samples. Therefore, ISPL can improve the performance of Eigen-Reptile.
	
	\section{Experimental Results and Discussion}
	\label{experiment}
	In our experiments, we aim to (1) compare different update directions of meta-parameters, (2) evaluate the effectiveness of Eigen-Reptile to alleviate overfitting sampling and label noise, (3) test the robustness of Eigen-Reptile to some hyperparameters, (4) evaluate the improvement of ISPL to Eigen-Reptile in the presence of noisy labels. All experiments run on a 2080 Ti. 
	
	\begin{table*}[t] 
		\centering
		\caption{ Accuracy of FSL on Mini-Imagenet N-way K-shot. The $\pm$ shows $95\%$ confidence interval over tasks. The number in $(\cdot)$ denotes the number of filters.} 
		\begin{tabular}{ccc} 
			\toprule 
			Algorithm & 5-way 1-shot & 5-way 5-shot \\
			\midrule 
			MAML \cite{finn2017model} 				& $48.70 \pm 1.84\%$  & $63.11 \pm 0.92\%$ \\
			FOML \cite{finn2017model}   		& $48.07 \pm 1.75\%$  & $63.15 \pm 0.91\%$ \\
			GNN \cite{gidaris2018dynamic}					& $50.30\%		$  & $66.40\%$\\
			TAML  \cite{jamal2019task}  	& $51.77 \pm 1.86\%$  & $65.60 \pm 0.93\%$ \\
			Meta-dropout \cite{lee2019meta}		&$51.93\pm0.67\%$		&$67.42\pm0.52\%$\\
			Warp-MAML \cite{flennerhag2020meta} &$52.30 \pm  0.80 \%$	& $68.4 \pm 0.60 \%$ \\
			MC (128) \cite{park2020meta}		& $\bm{54.08 \pm 0.93\%}$  & $67.99 \pm 0.73\%$ \\
			sparse-MAML \cite{von2021learning} & $51.04\pm0.59\%$ & $68.05\pm0.84\%$ \\
			MeTAL \cite{baik2021meta} & $52.63 \pm 0.37\%$ & $\bm{70.52 \pm 0.29\%}$\\
			MixtFSL \cite{afrasiyabi2021mixture} & $52.82 \pm0.63\%$ & $\bm{70.67 \pm0.57\%}$\\
			\midrule 
			Reptile (32) \cite{nichol2018first} 				& $49.97 \pm 0.32\%$  & $65.99 \pm 0.58\%$ \\
			Eigen-Reptile (32) 			& $51.80 \pm 0.90\%$  & $68.10\pm 0.50 \%$\\ 
			Eigen-Reptile (64)				&$\bm{ 53.25 \pm 0.45\%}$ &$ \bm{69.85\pm 0.85  \%}$\\
			\bottomrule
			\label{table1} 
			\vspace{-15pt}
		\end{tabular} 
	\end{table*}
	\subsection{Meta-learning with the Best Update Direction on Clean Dataset}
	
	In this experiment, we try to compare main direction with other update directions and evaluate the effectiveness of Eigen-Reptile to alleviate overfitting sampling noise by the 1D sine wave $K$-shot regression problem \cite{nichol2018first}. 
	Each task is defined by a sine curve $y(x) = A sin(x+b)$, where the amplitude $A \sim U ([0.1, 5.0])$ and phase $b \sim U ([0, 2\pi])$. The amplitude $A$ and phase $b$ are varied between tasks. The goal of each task is to fit a sine curve with the data points sampled from the corresponding $y(x)$. We calculate the loss in $\ell_2$ using 50 equally-spaced points from the whole interval $[-5.0,5.0]$ for each task. The loss is
	\begin{align}
	\int_{-5.0}^{5.0}  \parallel y(x)-\widehat{y}(x) \parallel^2 dx
	\end{align}
	where $\widehat{y}(x)$ is the predicted function that start from the initialization learned by meta-learner.
	
	The $K$-shot regression task fits a selected sine curve through $K$ points, here $K=10$. For the regressor, we use a small neural network, which is the same as \cite{nichol2018first}, except that the activation functions are Tanh. Specifically, the small network includes an input layer of size 1, followed by two hidden layers of size 64, and then an output layer of size 1. All meta-learners use the same regressor and are trained for 30000 iterations with inner loop steps 5, batch size 10, and a fixed inner loop learning rate of 0.02.
	\begin{figure}[h]
		\centering
		\includegraphics[width=0.35\textwidth]{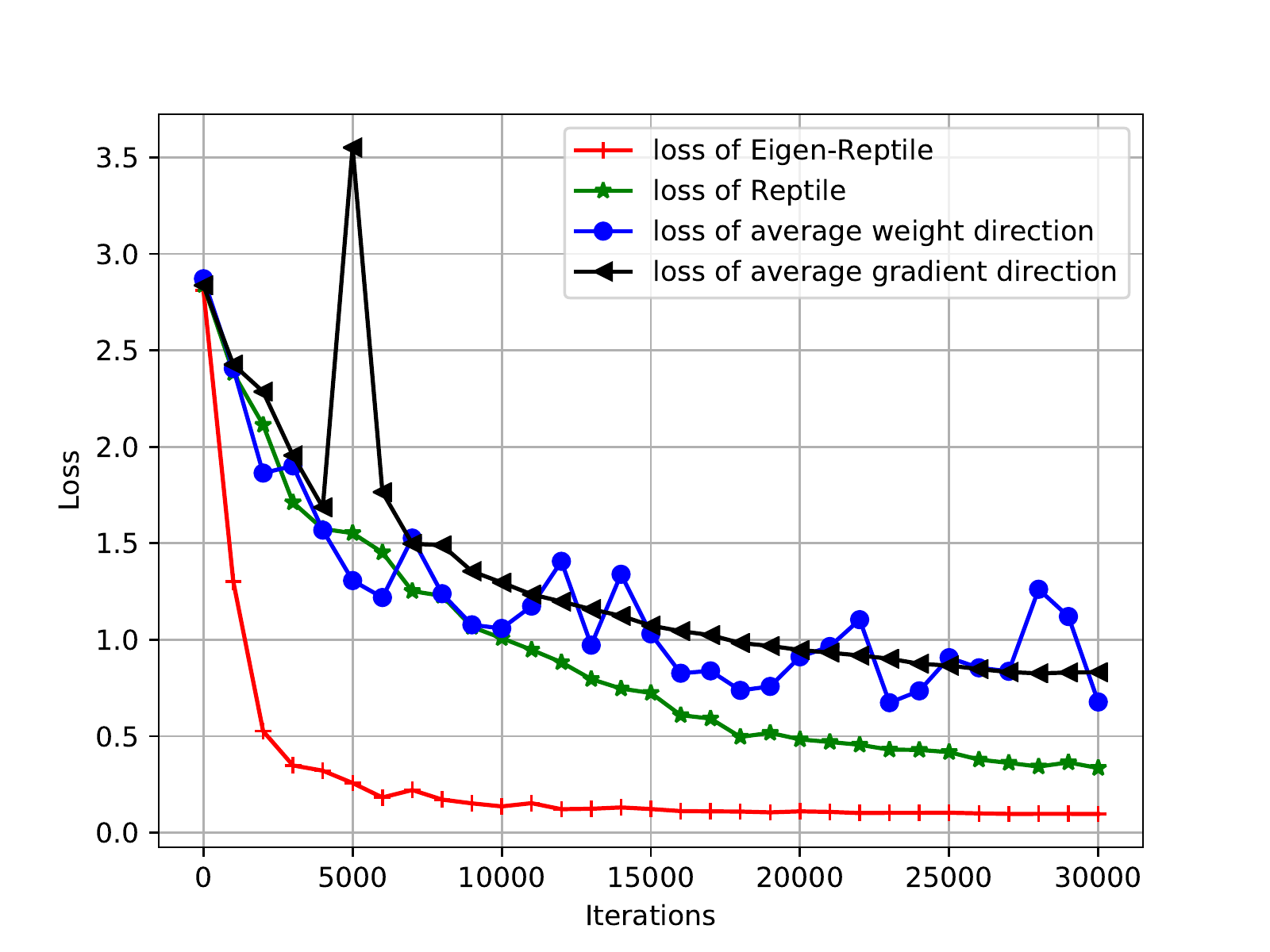}
		\vspace{-5pt}
		\caption{ Loss of different update direction. }
		\label{10-shot regression}
		\vspace{-5pt}
	\end{figure}
	
	We report the results of Reptile and Eigen-Reptile in \cref{img4}. It can be seen that the curve fitted by Eigen-Reptile is closer to the true green curve, which shows that Eigen-Reptile performs better. According to \cite{jamal2019task}, the initial model with a larger entropy before adapting to new tasks would better alleviate meta-overfitting. As shown in \cref{img4}, from 1 to 30000 iterations, Eigen-Reptile is more generalizable than Reptile as the initial blue line of Eigen-Reptile is closer to a straight line, which shows that the initialization learned by Eigen-Reptile is less affected by overfitting. Furthermore, \cref{10-shot regression} shows that update meta-parameters by the main direction converges faster and gets a lower loss than other update directions, such as the average of gradient direction and average of task-specific weights direction.

	\begin{table}[t] 
		\centering
		\caption{\label{tab:test} Few Shot Classification on CIFAR-FS N-way K-shot accuracy. The $\pm$ shows $95\%$ confidence interval over tasks.} 
		\begin{tabular}{ccc} 
			\toprule 
			Algorithm  &5-way 1-shot & 5-way 5-shot \\
			\midrule 
			MAML      & $58.90\pm 1.90 \% $& $71.50 \pm 1.00 \% $\\
			PROTO NET  	& $55.50 \pm 0.70\%$ & $72.00 \pm 0.60\%$ \\
			GNN			& ${61.90\%}$  & $75.30\%$ \\
			ECM	& ${55.14 \pm 0.48\%}$  & $71.66 \pm 0.39\%$\\
			\midrule 
			Reptile     &$58.30 \pm 1.20 \%$  & $75.45\pm 0.55\%$ \\
			Eigen-Reptile 			& $\bm{61.90\pm1.40\%}$  & $\bm{78.30\pm 0.50\%}$\\ 
			\bottomrule
			\label{Cifar-FS}
			\vspace{-23pt}
		\end{tabular} 
	\end{table}
	\subsection{Alleviate Overfitting Sampling Noise on Clean Dataset}
	We verify the effectiveness of Eigen-Reptile alleviate overfitting sampling noise on two clean few-shot classification datasets Mini-Imagenet \cite{vinyals2016matching} and CIFAR-FS \cite{bertinetto2018meta}. 
	\label{5.2}
	
	The Mini-Imagenet dataset contains 100 classes, each with 600 images. We follow \cite{ravi2016optimization} to divide the dataset into three disjoint subsets: meta-training set, meta-validation set, and meta-testing set with 64 classes, 16 classes, and 20 classes, respectively. We follow the few-shot learning protocols from prior work \cite{vinyals2016matching}, except that the number of the meta-training shot is 15 as \cite{lee2019meta,cao2019theoretical}, which is still much smaller than the number of samples required by traditional tasks. Moreover, we run our algorithm on the dataset for the different number of test shots and compare our results to other meta-learning algorithms. What needs to be reminded is that approaches that use deeper, residual networks or pretrained all-way classifier on meta-training set can achieve higher accuracies \cite{gidaris2018dynamic,yang2020dpgn,hu2020empirical}. So for a fair comparison, we only compare algorithms that use convolutional networks without a pretrained model as Reptile does. Specifically, 
	our model follows \cite{nichol2018first}, which has 4 modules with a $3 \times 3$ convolutions and 32 filters, $2 \times 2$ max-pooling etc. The images are downsampled to $84 \times 84$, and the loss function is the cross-entropy error. We use Adam with $\beta_1=0$ in the inner loop. Our model is trained for 100000 iterations with a fixed inner loop learning rate of 0.0005, 7 inner-loop steps and batch size 10.

	\begin{table*}[t]
		\centering
		\caption{Average test accuracy of 5-way 1-shot on the Mini-Imagenet with label noise. S and AS denotes symmetric and asymmetric noise, respectively. All methods are trained with early stopping to against noisy labels \cite{li2020gradient}, especially when $p=0.5$, the results of 20000 or more iterations for Reptile are only $20\%$ that equivalent to random guessing. Besides, for a fair comparison, we force all methods to get similar results when $p = 0$ to compare the robustness when $p$ is higher. Therefore, the reported results are lower than that of \cref{table1}.}
		\label{table2}
		\begin{tabular}{cccccccccccc}
			\toprule
			\multirow{2}{*}{Algorithm} & \multirow{2}{*}{$p=0.0$}& \multicolumn{2}{c}{$p=0.1$}& \multicolumn{2}{c}{$p=0.2$} & \multicolumn{2}{c}{$p=0.5$}\\
			\cmidrule(r){3-4} \cmidrule(r){5-6} \cmidrule(r){7-8}
			&       & S
			&  AS    &  S
			&  AS    &  S
			&AS		\\
			\midrule
			MeTAL \cite{baik2021meta}            &${47.80\%}$     &$44.64\%$         &$45.20\%$           &$39.80\%$          & $40.42\%$           & $27.05\%$    & $31.76\%$    \\
			Reptile \cite{nichol2018first}             &${47.64\%}$     &$46.08\%$         &$47.30\%$           &$43.49\%$          & $45.51\%$           & $23.33\%$    & $42.03\%$    \\
			Reptile+ISPL            &$47.23\%$    &$46.50\%$& $47.00\%$                      &$43.70\%$          & $45.42\%$           & $21.83\%$              &$41.09\%$ \\
			Eigen-Reptile            &$\bm{47.87\%}$     &$47.18\%$ &${47.42\%}$            &$45.01\%$          &$ 46.50\%$           & $27.23\%$       &$42.29\%$      \\
			Eigen-Reptile+ISPL             &$47.26\%$   &$\bm{47.20\%}$&$47.24\%$                    &$\bm{45.49\%}$          &$\bm{46.83\%}$           & $\bm{28.68\%}$         &$\bm{43.71\%}$    \\
			\bottomrule
			\vspace{-15pt}
		\end{tabular}
	\end{table*}
	
	The results of Eigen-Reptile and other meta-learning approaches are summarized in \cref{table1}. 
	The proposed Eigen-Reptile (64 filters) achieves highly competitive performance compared with other algorithms for 5-shot and 1-shot classification problems, respectively. More specifically, for 1-shot, the results of MC are similar to that of Eigen-Reptile. However, as a second-order optimization algorithm, the computational cost of MC is much higher than that of Eigen-Reptile (as shown in \textbf{\textit{Appendix}} \ref{AppendC}, Eigen-Reptile is a first-order algorithm like Reptile).
	As for 5-shot, the results of MeTAL and MixtFSL are slightly higher than that of Eigen-Reptile, while Eigen-Reptile performs better on the 1-shot task.
	Furthermore, the results of Eigen-Reptile with 32 filters are much better than that of Reptile with 32 filters for each task. Compared with Reptile, Eigen-Reptile uses the main direction to update the meta-parameters to alleviate the meta-overfitting caused by sampling noise. More importantly, Eigen-Reptile outperforms the state-of-the-art meta-overfitting preventing method Meta-dropout \cite{lee2019meta}, which is based on regularization. This result shows the effectiveness of addressing the meta-overfitting problem from the perspective of alleviating gradient noise. 
	
	The results of CIFAR-FS with new baseline ECM	\cite{ravichandran2019few} and PROTO Nets \cite{snell2017prototypical} are shown in Table \ref{Cifar-FS}. The settings of Eigen-Reptile in this experiment are the same as that of Mini-Imagenet experiments. 
	Moreover, we do not compare algorithms with additional tricks, such as higher way. It can be seen from Table \ref{Cifar-FS}, the performance of Eigen-Reptile is still far better than Reptile without any hyperparameter adjustment.

	We follow \cite{lee2019meta,cao2019theoretical} to vary the number of inner-loops and the number of corresponding	training shots to show the robustness of Eigen-Reptile on the 5-way 5-shot problem in \textbf{\textit{Appendix}} \ref{AppendG}, and results show that after the number of inner-loops $i$ reaches 7 with 15 train shots, the test accuracy tends to be stable.

	\subsection{Alleviate Overfitting Label Noise on Corrupted Dataset}
	\label{5.3}
	We conduct the 5-way 1-shot experiment with noisy labels generated by corrupting the original labels of Mini-Imagenet. There are symmetric label noise and asymmetric label noise in this experiment. For symmetric label noise, correct labels are flipped to other labels with equal probability, i.e., in the case of symmetric noise of ratio $p$, a sample retains the correct label with probability $1-p$, and it becomes some other label with probability $p/(N-1)$. On the other hand, for asymmetric label noise, we randomly flip the labels of one class to the labels of another fixed class with probability $p$.
	
    All meta-learners with 32 filters are trained with early stopping to against noisy labels \cite{li2020gradient} and get similar results when $p=0$. This experiment examines the robustness of different methods when facing different $p$ (i.e., when $p \neq 0$, the closer the results are to that of $p=0$, the better the method's robustness).
	The sample selection parameter $\gamma$ of ISPL is 10 that decreases by 0.6 every 1000 iterations, and the number of prior models is 2. Moreover, we only introduce noise during meta-training, where the train shot is 30. 
	
	As shown in \cref{table2}, for symmetric label noise, with the increase of the ratio $p$, the performance of Reptile decreases rapidly. When $p=0.5$, the initialization learned by Reptile can hardly meet the requirements of quickly adapting to new tasks with few samples. On the other hand, Eigen-Reptile is less affected by symmetric label noise, especially when the noise ratio is high, i.e., $p=0.5$. 
	As for asymmetric label noise,  meta-learners are trained on tasks with the same noise transition matrix, which allows meta-learners to learn more useful information, so the results are higher than that with symmetric noise. Like the results of symmetric label noise, Eigen-Reptile outperforms Reptile in all tasks.

	Eigen-Reptile+ISPL achieves better results than that of Eigen-Reptile when $p\not=0$. Specifically, ISPL plays a more significant role when $p$ is higher. However, when $p = 0$, ISPL harms Eigen-Reptile, as ISPL only discards correct samples. These results corresponding to the conclusion in \textbf{\textit{Appendix}} \ref{AppendD}, with the increase of noise ratio, the eigenvector will gradually become invalid, which demonstrates the effectiveness of ISPL and verify the idea that ISPL collaborates with Eigen-Reptile by providing improved main direction.
	In addition, ISPL does not significantly improve or even degrades the results of Reptile. This is because too many high-loss samples are removed, causing Reptile to fail to converge quickly with the same number of iterations.
	
	These experimental results show that Eigen-Reptile and ISPL can effectively separate noisy information, thereby alleviating overfitting label noise. 
	
	\section{Conclusion}
	This paper proposes a gradient-based meta-learning algorithm Eigen-Reptile. It updates the meta-parameters through the main direction, proven by theory and experiments to effectively alleviate the overfitting on sampling and label noise. Furthermore, to get closer to real-world situations, we introduce noisy labels into the meta-training dataset. The proposed ISPL constructs prior models to select samples for Eigen-Reptile to get a more accurate main direction.

	\section*{Acknowledgements}
	This work has been supported in part by National Key Research and Development Program of China (2018AAA0101900), Zhejiang NSF (LR21F020004), Alibaba-Zhejiang University Joint Research Institute of Frontier Technologies, Key Research and Development Program of Zhejiang Province, China (No. 2021C01013), Chinese Knowledge Center of Engineering Science and Technology (CKCEST).
	\nocite{langley00}
	
	\bibliography{example_paper}
	\bibliographystyle{icml2022}

	\newpage
	\appendix
	\onecolumn
	
	\section{PSEUDO-CODE }
	\label{AppendA}
	
	\begin{algorithm}[!h]
		\caption{Introspective Self-paced Learning}
		\textbf{Input}:  Dataset $D_{train}$, initialization $\boldsymbol{\phi}$, batch size $b$, selection parameter $\gamma$, attenuation coefficient $\mu$, the number of prior models $Q$.
		\begin{algorithmic}[1]
			\STATE Initialize network parameters $\boldsymbol{\phi}^{*} = \boldsymbol{\phi}$ for a sampled task
			\FOR{$j=1,2,3,\cdots,Q$}
			\STATE Sample examples $D_j$ from $D_{train}$ for training $model_j$, $\boldsymbol{\phi}_j=U^{m}(D_j,\boldsymbol{\phi}^*)$
			\ENDFOR
			\STATE Train task-specific parameters: 
			\FOR{$i = 1,2,3,\cdots,n$}
			\STATE Compute hidden variable vector $\boldsymbol{v}$:
			\STATE 	$v=  \arg \min _{\mathbf{v}} v_q\sum_{q=1}^{b}L_q -\gamma \sum_{q=1}^{b}v_q  $, \\ $\quad$ where $L_q=\frac{1}{Q}\sum_{j=1}^{Q} L_j\left(x_{q}, y_{q}, \boldsymbol{\phi}_j\right) $ 
			\STATE Update task-specific parameters $\phi^{*}$:
			\STATE $\boldsymbol{\phi}^{*} =\arg \min _{\boldsymbol{\phi}^{*} } L_{ISPL}\left(\boldsymbol{\phi}^*,v\right)$
			\STATE $\gamma=\gamma- \mu$
			\ENDFOR
		\end{algorithmic}
		\label{A2}
	\end{algorithm}
	
	\section{Proof of Theorem 1}
	\label{AppendB}
	Gradient update always with gradient noise inserted at every iteration due to sampling noise, which caused Reptile, MAML, etc. tend to overfit the trained samples. In this section, we will prove that Eigen-Reptile can alleviate meta-overfitting by alleviating gradient noise. 
	
	\noindent\textbf{Theorem 1}
	\textit{	Assume that the gradient noise variable $x$ follows Gaussian distribution \cite{hu2017diffusion,jastrzkebski2017three,mandt2016variational}, $x\sim\mathrm{N}\left(0, \sigma^2\right)$. Furthermore, $x$ and neural network parameter variable are assumed to be uncorrelated. The observed covariance matrix $\boldsymbol{C}$ equals noiseless covariance matrix $\boldsymbol{C}_t$ plus gradient noise covariance matrix $\boldsymbol{C}_x$. Then, we get}
	\begin{equation}
	\begin{aligned}
	\boldsymbol{C} =\frac{1}{n-1} \boldsymbol{S} =
	\boldsymbol{C}_t+\boldsymbol{C}_x =\boldsymbol{P}_t(\Lambda_t  + \boldsymbol{\Lambda}_x )\boldsymbol{P}_t^\top \\
	=\boldsymbol{P}_t(\boldsymbol{\Lambda}_t  + \sigma^2 \boldsymbol{I} )\boldsymbol{P}_t^\top=\boldsymbol{P}_t\Lambda \boldsymbol{P}_t^\top=\boldsymbol{P}\Lambda \boldsymbol{P}^\top
	\label{ET1}
	\end{aligned}
	\end{equation}
	\textit{where $\boldsymbol{P}_t$ and $\boldsymbol{P}$ are the orthonormal eigenvector matrices of $\boldsymbol{C}_t$ and $\boldsymbol{C}$ respectively, $\boldsymbol{\Lambda}_t$ and $\boldsymbol{\Lambda}$ are the corresponding diagonal eigenvalue matrices, and $\boldsymbol{I}$ is an identity matrix. It can be seen from \cref{ET1} that $\boldsymbol{C}$ and $\boldsymbol{C}_t$ has the same eigenvectors. }

	\begin{proof}	
		In the following proof, we assume that the probability density function of gradient noise variable $x$ follows Gaussian distribution, $x\sim\mathrm{N}\left(0, \sigma^2\right)$. Treat the parameters in the neural network as variables, and the parameters obtained by each gradient update as samples.  Furthermore, gradient noise and neural network parameters are assumed to be uncorrelated.
		
		For observed parameter matrix $\boldsymbol{W} \in R^{d \times n}$, there are $n$ samples, let $\boldsymbol{W}_{i,:} \in R^{1\times n}$ be the observed values of the i-th variable $\boldsymbol{W}_i$, and $\boldsymbol{W}=[\boldsymbol{W}_{1,:}^\top,\cdots,\boldsymbol{W}_{i,:}^\top,\cdots,\boldsymbol{W}_{d,:}^\top]^\top$. Similarly, we denote the noiseless parameter matrix by $\boldsymbol{W}^{t}=[(\boldsymbol{W}_{1,:}^t)^\top,\cdots,(\boldsymbol{W}_{i,:}^t)^\top,\cdots,(\boldsymbol{W}_{d,:}^t)^\top]^\top$, 
		and 
		\begin{align}
		\boldsymbol{W}=\boldsymbol{W}^t+\boldsymbol{X}
		\end{align} 
		Where $\boldsymbol{X}=[\boldsymbol{X}_{1,:}^\top,\cdots,\boldsymbol{X}_{i,:}^\top,\cdots,\boldsymbol{X}_{d,:}^\top]^\top$ is the dataset of noise variables. Then, centralize each variable by
		\begin{align}
		\overline{\boldsymbol{W}}_k=\boldsymbol{W}_k-\frac{1}{n}\sum_{i=1}^{n}{\boldsymbol{W}_{k,:}(i)}
		\end{align}
		So we get $\overline{\boldsymbol{W}}=[\overline{\boldsymbol{W}}_1^\top,\cdots,\overline{\boldsymbol{W}}_d^\top]^\top$.
		Suppose $\boldsymbol{W}^t$ is also centralized by the same way and get $\overline{\boldsymbol{W}^t}=[\overline{\boldsymbol{W}^t}_1^\top,\cdots,\overline{\boldsymbol{W}^t}_d^\top]^\top$. Then, we have:
		\begin{align}
		\overline{\boldsymbol{W}}= \overline{\boldsymbol{W}^t}+\boldsymbol{X}
		\end{align}
		Computing the covariance matrix of $\overline{\boldsymbol{W}}$:
		\begin{align}
		\begin{split}
		\boldsymbol{C}&= \frac{1}{n}\overline{\boldsymbol{W}} \overline{\boldsymbol{W}}^\top \\
		&=\frac{1}{n}(\overline{\boldsymbol{W}^t}+\boldsymbol{X})(\overline{\boldsymbol{W}^t}^\top+\boldsymbol{X}^\top) \\
		&=\frac{1}{n} (\overline{\boldsymbol{W}^t}\overline{\boldsymbol{W}^t}^\top +\overline{\boldsymbol{W}^t} \boldsymbol{X}^\top + \boldsymbol{X}\overline{\boldsymbol{W}^t}^\top + \boldsymbol{X}\boldsymbol{X}^\top)
		\end{split}
		\end{align}
		Since $\overline{\boldsymbol{W}^t}$ and $\boldsymbol{X}$ are uncorrelated, $\overline{\boldsymbol{W}^t} \boldsymbol{X}^\top$ and $\boldsymbol{X}\overline{\boldsymbol{W}^t}^\top $ are approximately zero matrices. Thus:
		\begin{align}
		\boldsymbol{C} \approx \frac{1}{n}(\overline{\boldsymbol{W}^t} \overline{\boldsymbol{W}^t}^\top + \boldsymbol{X}\boldsymbol{X}^\top) = \boldsymbol{C}_t+\boldsymbol{C}_x
		\end{align}
		The component $\boldsymbol{C}_x(i,j)$ is the correlation between $\boldsymbol{X}_i$ and $\boldsymbol{X}_j$ which corresponds to the i-th and j-th rows of $\boldsymbol{X}$. As the two noise variables are not related to each other, if $i\ne j$, then $\boldsymbol{C}_x(i,j)=0$. So $\boldsymbol{C}_x \in R^{d\times d}$ is a diagonal matrix with diagonal elements $\sigma^2$.
		Decompose $\boldsymbol{C}_t$ as:
		\begin{align}
		\boldsymbol{C}_t=\boldsymbol{P}_t\boldsymbol{\Lambda}_t \boldsymbol{P}_t^\top
		\end{align}
		where $\boldsymbol{P}_t$ is the noiseless orthonormal eigenvector matrix and $\boldsymbol{\Lambda}_t$ is the noiseless diagonal eigenvalue matrix, then
		\begin{align}
		\boldsymbol{C}_x&= \boldsymbol{\Lambda}_x \boldsymbol{P}_t\boldsymbol{P}_t^\top =\boldsymbol{P}_t \boldsymbol{\Lambda}_x \boldsymbol{P}_t^\top =\boldsymbol{P}_t \boldsymbol{C}_x \boldsymbol{P}_t^\top
		\end{align}
		where $\boldsymbol{\Lambda}_x = \sigma^2 \boldsymbol{I}$, and $\boldsymbol{I}$ is the identity matrix.
		Thus,
		\begin{align}
		\begin{split}
		\boldsymbol{C}&=\boldsymbol{C}_t+\boldsymbol{C}_x \\
		&=\boldsymbol{P}_t\boldsymbol{\Lambda}_t \boldsymbol{P}_t^\top + \boldsymbol{P}_t \boldsymbol{\Lambda}_x \boldsymbol{P}_t^\top \\
		&=\boldsymbol{P}_t(\boldsymbol{\Lambda}_t  + \boldsymbol{\Lambda}_x )\boldsymbol{P}_t^\top \\
		&=\boldsymbol{P}_t \boldsymbol{\Lambda} \boldsymbol{P}_t^\top
		\label{22}
		\end{split}
		\end{align}
		where $\boldsymbol{\Lambda} = \boldsymbol{\Lambda}_t + \boldsymbol{\Lambda}_x$. It can be seen from  \cref{22} that $\boldsymbol{C}$ and $\boldsymbol{C}_t$ has the same eigenvector matrix. In other words, \textbf{eigenvector is not affected by gradient noise.}
		\label{proof1}
	\end{proof}	
	
	\section{Algorithm Complexity Analysis of Eigen-Reptile}
	\label{AppendC}
	As for the time complexity of Eigen-Reptile, the cost of single gradient descent in the inner-loop is $\mathcal{O}(d)$, where $d$ is the number of network parameters. The cost of the scatter/covariance matrix computations is 
	$\mathcal{O}(n^2d)$, where $n$ is the number of inner-loop. Moreover, the worst-case complexity of computing eigenvalue decomposition is $\mathcal{O}(n^3)$. Finally, the computational complexity of restoring eigenvector is $\mathcal{O}(nd)$. We set the maximal number of outer-loop to $T$. Hence the time complexity of Eigen-Reptile is 
	\begin{equation}
	\begin{aligned}
	\mathcal{O}(T(n^3+n^2d&+nd+nd))= \mathcal{O}(Td)
	\end{aligned}
	\end{equation}
	In FSL, $n$ is small (in this paper $n=7$ ), so the overall time complexity is $\bm{\mathcal{O}(Td)}$.
	
	As for Reptile, the time complexity is also $\mathcal{O}(Td)$, which means that the time complexity of both Reptile and Eigen-Reptile is much lower than the second-order optimization algorithms.
	
	As for spatial complexity, Eigen-Reptile needs to store a $d\times n$ matrix and a $n\times n$ matrix. The overall spatial complexity is $\bm{\mathcal{O}(d)}$, while the spatial complexity of Reptile is $\mathcal{O}(d)$, too.
	
	It can be seen that, compared to Reptile, Eigen-Reptile is the same in spatial complexity and time complexity. Still, the accuracy of Eigen-Reptile is much higher than that of Reptile.
	
	\section{ Effectiveness of Eigen-Reptile for Label Noise}
	\label{AppendD}
	Eigen-Reptile alleviates the overfitting on label noise by separating noisy information. More specifically, from the perspective of signal to noise ratio (SNR) :
	\begin{align}
	SNR= \frac{ \delta_{clean}^2}{\delta_{noise}^2}
	\end{align}
	where $\delta_{clean}^2$ is the variance of weight points introduced by clean data, and $\delta_{noise}^2$ is the variance of weight points introduced by corrupted data. \\
	Normally, SNR$\gg$1, which shows that the effective information is far more than noisy information. Specifically, in a task-specific training, weight matrix $\mathbf{W}$ is composed of all weight points, and we decompose it as:
	\begin{align}
	\mathbf{W}=\mathbf{U S V}^{T}=\sum_{k=1}^{r} \sigma_{k} \mathbf{u_k} \mathbf{v_k}^{T}
	\end{align}
	where $r$ is the rank of $W$, $u_k$ is the eigenvector of the covariance matrix $WW^{T}$, $v_k$ is the eigenvector of the covariance matrix $W^{T}W$. \\
	The singular values are ordered in decreasing magnitude as 
	$\sigma_{1} \geq \sigma_{2} \geq \cdots \geq \sigma_{r}$ which are the positive square roots of the eigenvalues of $WW^{T}$. In Eigen-Reptile, we only remain the eigenvector corresponding to the largest eigenvalue, e.g., $\sigma_{1}^2$, which can be viewed as the $\delta_{clean}^2$ in SNR. In contrast, the noisy information $\delta_{noise}^2$ is removed by omitting the low singular values. 
	However, with the increase of noise ratio, Eigen-Reptile is more and more likely to suffer from the problem that the influence of corrupted samples gradually dominates the main update direction.
	
	\section{Proof of Theorem 2}
	\label{AppendE}
	In this section, we will prove that discarding high loss samples will result in a more accurate main direction when there are noisy labels.
	
	\noindent\textbf{Theorem 2} 
	\textit{	Let $\boldsymbol{W}_o$ be the parameter matrix only generated by the corrupted samples.
		Compute the eigenvalues and eigenvectors of the observed expected parameter matrix}
	\begin{equation}
	\begin{aligned}
	\frac{1}{\lambda} \mathbb{E}(\boldsymbol{C}_{tr})\boldsymbol{e}&=\boldsymbol{P}_{o}(\boldsymbol{I}-\frac{\boldsymbol{\Lambda}_{o}}{\lambda})\boldsymbol{P}_{o}^\top \boldsymbol{e} \\
	&\approx  \boldsymbol{P}_{o}(\boldsymbol{I}-\frac{\lambda_{o}}{\lambda}\boldsymbol{I})\boldsymbol{P}_{o}^\top \boldsymbol{e}>\boldsymbol{P}_{o}(\boldsymbol{I}-\frac{\lambda_{o}-\xi}{\lambda-\xi}\boldsymbol{I})\boldsymbol{P}_{o}^\top \boldsymbol{e}
	\label{APP12}
	\end{aligned}
	\end{equation}
	\textit{where $\boldsymbol{C}_{tr}$ is the covariance matrix generated by clean samples, $\lambda$ is the observed largest eigenvalue, $\lambda_{o}$ is the largest eigenvalue in the corrupted diagonal eigenvalue matrix $\boldsymbol{\Lambda}_{o}$, $\boldsymbol{P}_o$ is the orthonormal eigenvector matrix of corrupted covariance matrix. According to \cref{APP12}, if $\lambda_{o} /\lambda$ is smaller, the observed eigenvector $e$ is more accurate. Assume that the discarded high loss samples have the same contributions $\xi$ to $\lambda$ and $\lambda_{o}$, representing the observed and corrupted main directional variance, respectively. Note that these two kinds of data have the same effect on the gradient updating of the model, as they all generate high loss (neither new clean samples from new tasks nor corrupted samples are not familiar to the model). 
		Furthermore, it is easy to find that $(\lambda_{o}-\xi) /(\lambda-\xi)$ is smaller than $\lambda_{o} /\lambda$. }
	
	\begin{proof}
		Here, we use $\boldsymbol{w}$ to represent the parameter point obtained after a gradient update. For convenience, let $\boldsymbol{w}$ be generated by a single sample, $\boldsymbol{w}\in R^{d\times 1}$. Then the parameter matrix can be obtained, 
		\begin{equation}
		\boldsymbol{W}=
		\begin{bmatrix} 
		\boldsymbol{w}_1^{tr}  & \boldsymbol{w}_2^{tr} & \cdots & \boldsymbol{w}_1^o & \cdots & \boldsymbol{w}_m^o & \cdots & \boldsymbol{w}_n^{tr}\\ 
		\end{bmatrix}
		\end{equation}
		where $\boldsymbol{w}^o$ represents the parameters generated by the corrupted sample, and $\boldsymbol{w}^{tr}$ represents the parameters generated by the true sample. Furthermore, there are $n$ parameter points generated by $n$ samples. Moreover, there are $m$ corrupted parameter points generated by $m$ corrupted samples. Mean centering $\boldsymbol{W}$, and show the observed covariance matrix $\boldsymbol{C}$ as
		\begin{align}
		\begin{split}
		\boldsymbol{C}&=\frac{1}{n}\boldsymbol{W}\boldsymbol{W}^\top \\
		&=\frac{1}{n}
		\begin{bmatrix} 
		\boldsymbol{w}_1^{tr}  & \boldsymbol{w}_2^{tr} & \cdots & \boldsymbol{w}_n^{tr}\\ 
		\end{bmatrix}
		\begin{bmatrix} 
		(\boldsymbol{w}_1^{tr})^\top \\  (\boldsymbol{w}_2^{tr})^\top \\ \vdots \\ (\boldsymbol{w}_n^{tr})^\top\\ 
		\end{bmatrix}\\
		&=\frac{1}{n}(\boldsymbol{w}_1^{tr}(\boldsymbol{w}_1^{tr})^\top+\cdots+\boldsymbol{w}_1^o(\boldsymbol{w}_1^o)^\top+\cdots+\boldsymbol{w}_m^o(\boldsymbol{w}_m^o)^\top\\
		&\quad+\cdots+\boldsymbol{w}_n^{tr}(\boldsymbol{w}_n^{tr})^\top)
		\label{29}
		\end{split}
		\end{align}
		It can be seen from the decomposition of $\boldsymbol{C}$ that the required eigenvector is related to the parameters obtained from the true samples and the parameters obtained from the noisy samples. For a single parameter point
		\begin{align}
		\begin{split}
		\boldsymbol{w}\boldsymbol{w}^\top&=
		\begin{bmatrix} 
		a_1  \\ \vdots \\ a_i \\ \vdots \\ a_d
		\end{bmatrix}
		\begin{bmatrix} 
		a_1  & \cdots & a_i & \cdots & a_d
		\end{bmatrix}	\\
		&=
		\begin{bmatrix}
		a_{1}^2 & a_{1}a_2 & \cdots & a_1a_{d}\\
		a_{2}a_1 & a_{2}^2 & \cdots & a_2a_{d}\\
		\vdots & \vdots & \ddots & \vdots\\
		a_{d}a_1 & a_{d}a_2 & \cdots & a_{d}^2\\
		\end{bmatrix}
		\end{split}
		\end{align}
		As we discard all high loss samples that make the model parameters change significantly, and the randomly generated noisy labels may cause the gradient to move in any direction, we assume that the variance of corrupted parameter point variables is $\delta$. Compute the expectations of all variables in the corrupted parameter point
		\begin{equation}
		\begin{aligned}
		&\mathbb{E}(\boldsymbol{w}\boldsymbol{w}^\top) =\\
		&
		\begin{bmatrix}
		\delta_1^2+	\mathbb{E}(a_1)^2 & 	\mathbb{E}(a_1a_2) & \cdots &  	\mathbb{E}(a_1a_d)\\
		\mathbb{E}(a_2a_1) &  \delta_2^2+	\mathbb{E}(a_2)^2 & \cdots &  	\mathbb{E}(a_2a_d)\\
		\vdots & \vdots & \ddots & \vdots\\
		\mathbb{E}(a_da_1) &  	\mathbb{E}(a_da_2) & \cdots &  \delta_d^2+	\mathbb{E}(a_d)^2\\
		\end{bmatrix}
		=
		\Omega
		\label{31}
		\end{aligned}
		\end{equation}
		Let the sum of all corrupted $\frac{1}{n}	\mathbb{E}(\boldsymbol{w}\boldsymbol{w}^\top)$ be $\boldsymbol{\Omega}_o$, then
		\begin{align}
		\begin{split}
		&\boldsymbol{\Omega}_o =\\
		&\frac{1}{n}
		\begin{bmatrix}
		m{{\tiny \delta^2}}+\sum_{j=1}^{m}{	\mathbb{E}(a_{j1})^2} &  \cdots & \sum_{j=1}^{m}{	\mathbb{E}(a_{j1}a_{jd})}\\
		\sum_{j=1}^{m}{	\mathbb{E}(a_{j2}a_{j1})}  & \cdots & \sum_{j=1}^{m}{	\mathbb{E}(a_{j2}a_{jd})}\\
		\vdots & \ddots & \vdots\\
		\sum_{j=1}^{m}{	\mathbb{E}(a_{jd}a_{j1})}  & \cdots &  m{\delta^2}+\sum_{j=1}^{m}{	\mathbb{E}(a_{jd})^2}
		\end{bmatrix}
		\label{14}
		\end{split}
		\end{align}
		And let the sum of all true $\frac{1}{n}\boldsymbol{w}\boldsymbol{w}^\top$ be $\boldsymbol{C}_{tr}$. So the expectation of $\boldsymbol{C}$ can be written as,
		\begin{align}
		\begin{split}
		\mathbb{E}(\boldsymbol{C})
		=	\mathbb{E}(\boldsymbol{C}_{tr})+\boldsymbol{\Omega}_o
		\end{split}
		\end{align}
		Treat eigenvector and eigenvalue as definite values, we get
		\begin{align}
		\begin{split}
		(\boldsymbol{\Omega}_o+	\mathbb{E}(\boldsymbol{C}_{tr}))\boldsymbol{e}&=\lambda \boldsymbol{e} \\
		\end{split}
		\end{align}
		where $\boldsymbol{e}$ is the observed eigenvector, $\lambda$ is the corresponding eigenvalue. Divide both sides of the equation by $\lambda$.
		\begin{align}
		\begin{split}
		\frac{1}{\lambda} 	\mathbb{E}(\boldsymbol{C}_{tr})\boldsymbol{e}&=(\boldsymbol{I}- \frac{1}{\lambda}\boldsymbol{\Omega}_o)\boldsymbol{e} \\
		&=\boldsymbol{P}_o(\boldsymbol{I}-\frac{1}{\lambda}\boldsymbol{\Lambda}_o)\boldsymbol{P}_o^\top \boldsymbol{e}\\
		&\approx \boldsymbol{P}_{o}(\boldsymbol{I}-\frac{\lambda_{o}}{\lambda}\boldsymbol{I})\boldsymbol{P}_{o}^\top \boldsymbol{e}
		\label{43}
		\end{split}
		\end{align}
		where $\lambda_{o}$ is the largest eigenvalue in the corrupted diagonal eigenvalue matrix $\boldsymbol{\Lambda}_{o}$, $\boldsymbol{P}_o$ is the orthonormal eigenvector matrix of $\boldsymbol{\Omega}_o$. According to \cref{43}, if $\lambda_{o} /\lambda$ is smaller, $\boldsymbol{e}$ is more accurate. Discard some samples with the largest losses, which may contain true samples and noisy samples.  Assume that the discarded high loss samples have the same contributions $\xi$ to $\lambda$ and $\lambda_{o}$, as these two kinds of data have the same effect on the gradient updating of the model. Compare the ratio of eigenvalues before and after discarding, get
		\begin{align}
		\begin{split}
		\underbrace{\frac{\lambda_{o}}{\lambda}}_{before} - \underbrace{\frac{\lambda_{o}-\xi}{\lambda-\xi}}_{after}
		= \frac{\xi(\lambda-\lambda_{o})}{\lambda(\lambda-\xi)}>0
		\label{36}
		\end{split}
		\end{align}
		Obviously, $\lambda>\lambda_{o}$, and if we don't discard all samples, then $\lambda>\xi$. So \cref{36}$>0$, which means discarding high loss samples could reduce $\lambda_{o} /\lambda$. \textbf{Therefore, discarding high loss samples can improve the accuracy of eigenvector in the presence of noisy labels.}
		
		For further analysis, we assume that any two variables are independently and identically distributed, the expectation of variable $a$, $\mathbb{E}(a)=\epsilon$. Thus,
		\begin{align}
		\frac{1}{\lambda}\boldsymbol{\Omega}_o=\frac{p}{\lambda}
		\begin{bmatrix}
		{\delta^2}+\epsilon^2 &  \cdots & \epsilon^2 \\
		\epsilon^2   & \cdots & \epsilon^2 \\
		\vdots & \ddots & \vdots\\
		\epsilon^2   & \cdots &  {\delta^2}+\epsilon^2
		\end{bmatrix}
		\label{simply}
		\end{align}
		where $p$ is the proportion of noisy labels, $np=m$. As can be seen from \cref{simply}, if $p\epsilon^2/\lambda \approx0$, then $\boldsymbol{\Omega}_o/{\lambda}$ is a diagonal matrix. According to proof. \ref{proof1}, the observed eigenvector $\boldsymbol{e}$ is unaffected by noisy labels with the corresponding eigenvalue $\frac{p(\delta^2+\epsilon^2)}{\lambda}$.
	\end{proof}

	\section{Hyperparameter Analysis}
	\label{AppendG}
	\begin{figure}[h]
		\centering
		\includegraphics[width=0.5\textwidth]{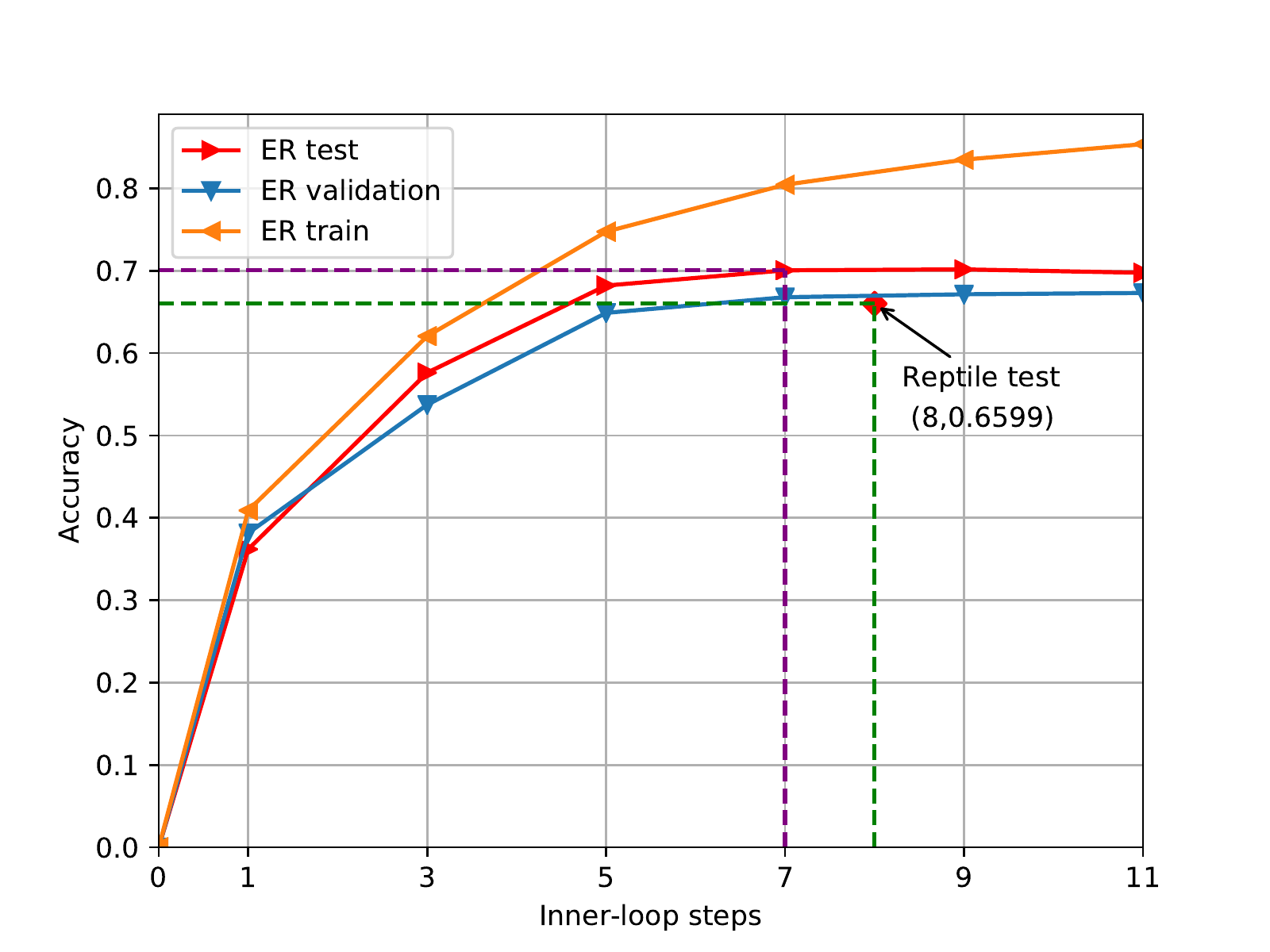}
		\caption{ The number of inner-loop and accuracy of 5-way 5-shot task on Mini-Imagenet.}
		\label{ha}
	\end{figure}
	As shown in \cref{ha}, after the number of inner-loops $i$ reaches 7, the test accuracy tends to be stable, which shows that changing the number of inner-loops within a specific range has little effect on Eigen-Reptile. That is, Eigen-Reptile is robust to this hyperparameter. As for train shot, to make the trained task-specific parameters as unbiased as possible, we specify train shot roughly satisfies $\lceil \frac{i\times batch\_size}{N} \rceil+1$, where $N$ is the number of classes. So when $i=7$, the number of train shots is 15.
	It is important to note that in our experiments, Reptile with the hyperparameters set by its authors, the number of inner-loops is 8, the number of train shots is 15, and the corresponding accuracy is $65.99\%$, which is much lower than the result of Eigen-Reptile.
	
	\section{Relationship between Eigen-Reptile and ISPL}
	Eigen-Reptile and ISPL are strongly correlated because of the following three aspects: (1) intuitively, with the increase of noise ratio, Eigen-Reptile is more likely to suffer from the problem that the influence of corrupted samples gradually dominates the main update direction. Therefore, ISPL is proposed to improve the accuracy of the main direction for Eigen-Reptile by introspective priors. (2) theoretically, Theorem \ref{TT2} shows that ISPL discards high loss samples to reduce $\lambda_o$ / $\lambda$ in Equation \ref{PP12}, which helps improve the accuracy of the observed eigenvector learned with corrupted samples. (3) empirically, in Table \ref{table2}, stable performance improvement of Eigen-Reptile + ISPL over Eigen-Reptile can be observed, and the performance gap becomes more significant when the noise ratio $p$ increases (which corresponds to the conclusion in \textbf{\textit{Appendix}} \ref{AppendD} - the high noise ratio is more likely to degrade the largest singular value and the corresponding eigenvector). The experimental results demonstrate the effectiveness of ISPL and verify the idea that ISPL collaborates with Eigen-Reptile by providing improved main direction (which would be more helpful as the noise ratio becomes larger).
\end{document}